\documentclass[10pt,twocolumn,letterpaper]{article}

\usepackage[pagenumbers]{cvpr} 
\usepackage[accsupp]{axessibility}  
\usepackage{url}            
\usepackage{booktabs}       
\usepackage{amsfonts}       
\usepackage{nicefrac}       
\usepackage{microtype}  
\usepackage{amsmath}
\usepackage{colortbl}
\usepackage{fancyhdr}
\usepackage{amssymb}
\usepackage{graphicx}
\usepackage{tabularx}
\usepackage{mathtools}
\usepackage{multirow}
\usepackage{arydshln}
\usepackage{stfloats}

\usepackage{placeins}

\definecolor{cvprblue}{rgb}{0.21,0.49,0.74}
\usepackage[pagebackref,breaklinks,colorlinks,allcolors=cvprblue]{hyperref}

\def\paperID{5573} 
\def\confName{CVPR}
\def\confYear{2025}

\title{TAMT: Temporal-Aware Model Tuning for
Cross-Domain \\ Few-Shot  Action Recognition}

\author{Yilong Wang$^{1,}$\thanks{Equal contribution.~~$\dag$Corresponding author: qlwang@tju.edu.cn.}~~~Zilin Gao$^{2, *}$~Qilong Wang$^{1, \dag}$ ~Zhaofeng Chen$^{3}$ Peihua Li$^{2}$~Qinghua Hu$^{1}$\\
\textsuperscript{1}Tianjin University \textsuperscript{2}Dalian University of Technology \textsuperscript{3}Yancheng Institute of Technology 
}

\begin{document}    
\maketitle

\definecolor{colorbaseline}{RGB}{224,238,238}
\definecolor{colorist}{RGB}{225, 240, 213}
\definecolor{colorpydist}{RGB}{255,214,165}
\definecolor{colorpydistlt}{RGB}{255,245,205}
\definecolor{colormtfan}{RGB}{242,216,137}
\definecolor{colorpink}{RGB}{255, 182, 193}
\definecolor{colorgrey}{RGB}{230, 230, 230}
\newcommand{\blu}[1]{\textcolor{blue}{#1}}
\newcommand{\pgrey}[1]{\colorbox{colorbaseline}{#1}}
\newcommand{\ssgap}[1]{\colorbox{colorist}{#1}}
\newcommand{\hlours}[1]{\colorbox{colorpydist}{#1}}
\newcommand{\ppink}[1]{\colorbox{colorpink}{#1}}
\newcommand{\psp}[1]{\colorbox{colorgrey}{#1}}

\begin{abstract}
Going beyond few-shot action recognition (FSAR), cross-domain FSAR (CDFSAR) has attracted recent research interests by solving the domain gap lying in source-to-target transfer learning. Existing CDFSAR methods mainly focus on joint training of source and target data to mitigate the side effect of domain gap. However, such kind of methods suffer from two limitations: First, pair-wise joint training requires retraining deep models in case of one source data and multiple target ones, which incurs heavy computation cost, especially for large source and small target data. Second, pre-trained models after joint training are adopted to target domain in a straightforward manner, hardly taking full potential of pre-trained models and then limiting recognition performance. To overcome above limitations, this paper proposes a simple yet effective baseline, namely Temporal-Aware Model Tuning (TAMT) for CDFSAR. Specifically, our TAMT involves a decoupled paradigm by performing pre-training on source data and fine-tuning target data, which avoids retraining for multiple target data with single source. To effectively and efficiently explore the potential of pre-trained models in transferring to target domain, our TAMT proposes a Hierarchical Temporal Tuning Network (HTTN), whose core involves local temporal-aware adapters (TAA) and a global temporal-aware moment tuning (GTMT). Particularly, TAA learns few parameters to recalibrate the intermediate features of frozen pre-trained models, enabling efficient adaptation to target domains. Furthermore, GTMT helps to generate powerful video representations, improving match performance on the target domain. Experiments on several widely used video benchmarks show our TAMT outperforms the recently proposed counterparts by 13\%$\sim$31\%, achieving new state-of-the-art CDFSAR results.
\end{abstract}    
\section{Introduction}
\label{sec:intro}
\begin{figure*}[tb]
  \centering
  \begin{subfigure}{0.63\linewidth}
  \centering
    \includegraphics[height=4.8cm]{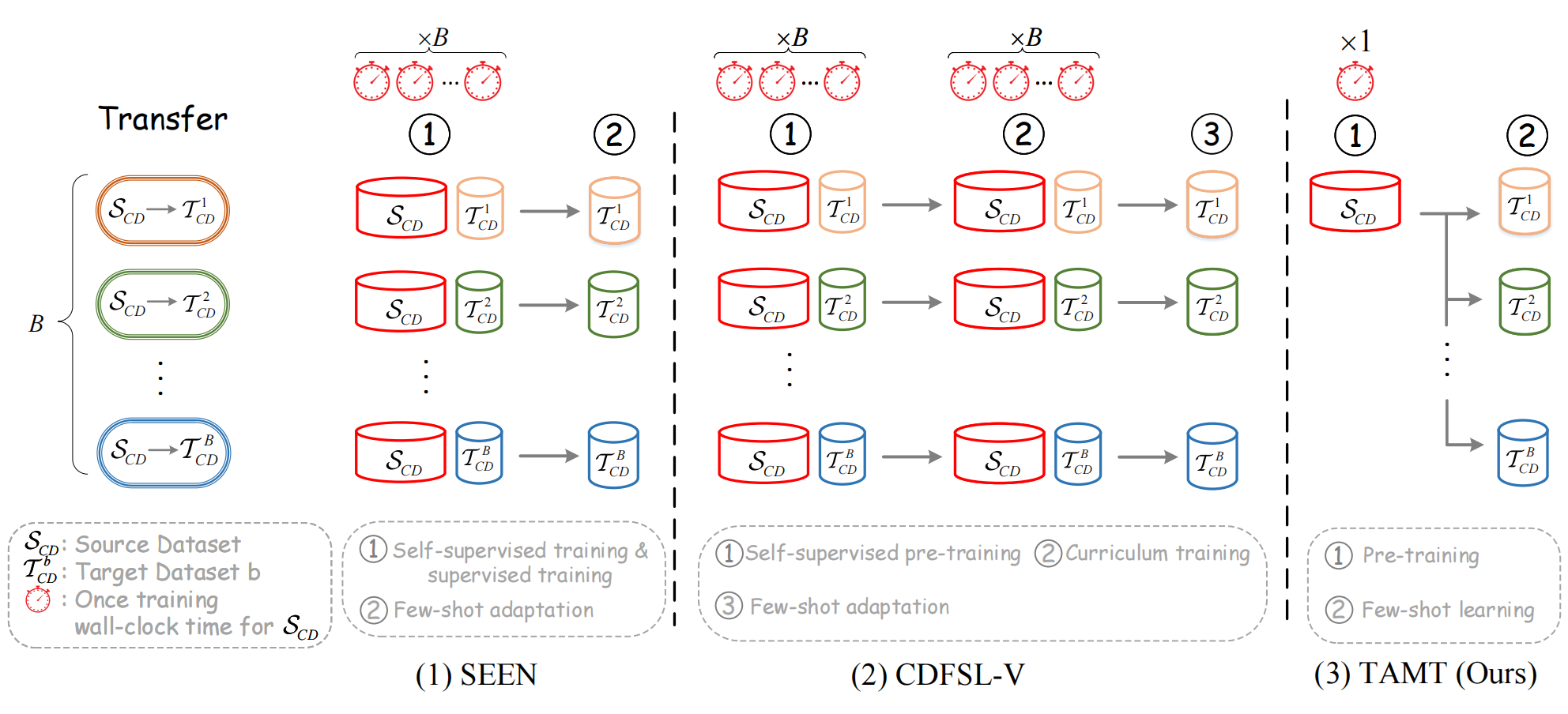}
    \caption{Different Training Paradigms}
    \label{fig:pipeline}
  \end{subfigure}
  \hfill
  \begin{subfigure}{0.36\linewidth}
  \centering
    \includegraphics[height=5cm]{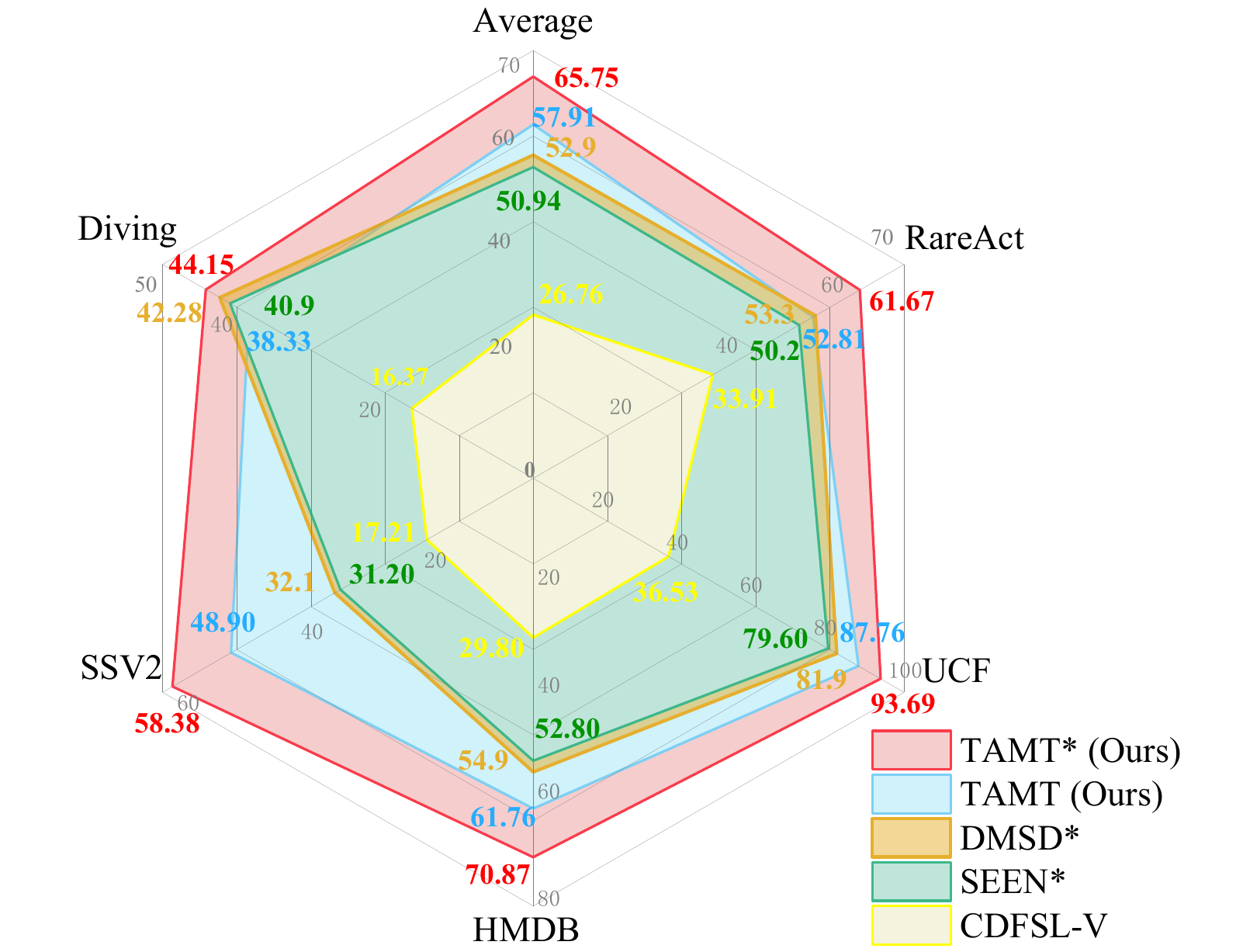}
    \caption{Performance of Existing CDFSAR Methods}
    \label{fig:radar}
  \end{subfigure}
  \caption{(a) Comparison of existing CDFSAR  methods in terms of training paradigm under the case of a single source data $\mathcal{S}_{CD}$ and $B$ target data $\{\mathcal{T}_{CD}^{1},\cdots,\mathcal{T}_{CD}^{B}\}$. (b) Comparison (\%) of existing CDFSAR methods with K-100 as source data. All results are conducted with $112\times 112$ resolution except methods marked by * ($224\times 224$ resolution)}
  \label{fig:1}
\end{figure*}

Few-shot action recognition (FSAR) aims to develop video recognition models with high generalization ability by using limited annotated samples, which has achieved remarkable progress with the rapid development of deep models and pre-training techniques~\cite{2020otam, wang2023molo, 2022Hybrid, GgHM_ICCV23, AMFTR_CVPR23, wang2023clip, wang2023few, wang2023few2, wang2023task, 2312.01083, Liu_2023_WACV}. Going beyond FSAR, cross-domain FSAR (CDFSAR) has been attracting recent research interests~\cite{WANG2023103737,10378593}, which focuses on transferring knowledge from the well-annotated source 
domain to target one with few annotated samples. Intuitively, the domain gap between source and target data will clearly impact the performance of transfer learning~\cite{Cross_domain_r1, Cross_domain_r2, Cross_domain_r3}. 

As a seminal work, SEEN~\cite{WANG2023103737} proposes a joint training paradigm to alleviate side effect of domain gap, where a parameter-shared model is trained on source data and target one with supervision learning and self-supervised contrastive learning objectives, respectively. After the joint training, a simple nearest neighbor classifier is straightforwardly used for inference in target domain. As a parallel work, CDFSL-V~\cite{10378593} proposes a two-stage joint training paradigm, where the model is first pre-trained on source and target data in a self-supervised manner, and then a curriculum learning is developed to further tune the model on source and target data. Subsequently, a few-shot classifier is fine-tuned on the annotated target data for inference. 

Although some advanced efforts are made~\cite{WANG2023103737,10378593}, they generally suffer from two limitations. First, both SEEN~\cite{WANG2023103737} and CDFSL-V~\cite{10378593} involve joint training paradigms. As illustrated in Fig.~\ref{fig:pipeline} (1) and (2), they require to retrain the models $B$ times, given a single source data $\mathcal{S}_{CD}$ and $B$ target data $\{\mathcal{T}_{CD}^{1},\cdots,\mathcal{T}_{CD}^{B}\}$ (a commonly used setting~\cite{WANG2023103737,10378593}). It potentially incurs heavy computation cost due to frequent retraining on source data $\mathcal{S}_{CD}$, especially for large $\mathcal{S}_{CD}$ and small $\mathcal{T}_{CD}^{b}$. Second, during the inference stage, pre-trained models are generally adopted to target domain in a straightforward manner, i.e., simple nearest neighbor classifier~\cite{WANG2023103737} or a fine-tuned classifier~\cite{10378593}. They hardly take full advantage of pre-trained models to dynamically fit target data with the frozen backbone, and so potentially limit the final recognition performance.

To address the above limitations, this paper proposes a simple yet effective baseline, namely Temporal-Aware Model Tuning (TAMT). Particularly, as shown in Fig.~\ref{fig:pipeline} (3), our TAMT involves a decoupled paradigm by pre-training the model on source data and subsequently fine-tuning it on target data. For model pre-training, we introduce a self-supervised followed by a supervised learning scheme to consider abilities of both generalization and semantic features extraction. As such, in the case of one source data and multiple target data, our TAMT only requires model pre-training one time, significantly decreasing training cost.

To explore the potential of pre-trained models in domain adaptation, our TAMT proposes a Hierarchical Temporal Tuning Network (HTTN), whose core involves local Temporal-Aware Adapters (TAA) and a Global Temporal-aware Moment Tuning (GTMT). Particularly, TAA introduces few learnable parameters to recalibrate a part of intermediate features outputted by frozen pre-training models, which helps adapt pre-training models to target data efficiently. By considering the significance of global representations in metric-based few-shot classification, our GTMT proposes to exploit spatio-temporal feature distribution approximated first- and second-order moments to generate powerful video representations. Particularly, GTMT presents an efficient long-short temporal covariance (ELSTC) to effectively compute second-order moments of spatio-temporal features. By equipping with TAA and GTMT, our HTTN dynamically adopts pre-trained models to target data in an effective and efficient way, clearly improving recognition performance. As shown in Fig.~\ref{fig:radar}, our proposed TAMT can bring significant performance gains over existing methods with lower training cost. To evaluate our TAMT, experiments are conducted on 
five source datasets (i.e., Kinetics-400 (K-400)~\cite{2017The}, Kinetics-100 (K-100)~\cite{2018Compound}, Something-Something V2 (SSV2)~\cite{2017The2}, Diving48 (Diving)~\cite{2018RESOUND} and UCF-101 (UCF)~\cite{2012UCF101}) and five target datasets (i.e., HMDB51 (HMDB)~\cite{2011HMDB}, SSV2, Diving, UCF-101 (UCF)~\cite{2012UCF101} and RareAct~\cite{2020RareAct}). The contributions of this work can be summarized as follows:
\begin{itemize}
\item [1)] In this paper, we propose a simple yet effective baseline for the cross-domain few-shot action recognition (CDFSAR) task, namely Temporal-Aware Model Tuning (TAMT). 
To our best knowledge, TAMT makes the first attempt to introduce a decoupled training paradigm for CDFSAR, effectively avoiding frequent retraining in the case of one source data and multiple target data.

\item [2)] Unlike previous CDFSAR  works, our TAMT pays more attention to effectively and efficiently adopting pre-trained models to target data. Particularly, a lightweight Hierarchical Temporal Tuning Network (HTTN) is proposed to recalibrate intermediate features and generate powerful video representations for the frozen pre-training models via local Temporal-Aware Adapters (TAA) and a Global Temporal-aware Moment Tuning (GTMT), respectively.

\item [3)] Extensive experiments are conducted on various video benchmarks, and the results show our TAMT significantly outperforms the recently proposed CDFSAR methods.
\end{itemize}
\newcommand{\tocheck}[1]{\textcolor{red}{#1}}
\section{Related Work}
\subsection{Few-Shot Action Recognition}
With the development of large video models and the insurmountable success of action recognition methods~\cite{arnab2021vivit, 9970717, 2021Semantic,2022Multiview,wang2022internvideo, videomaev2, 10489992, 0Rethinking}, few-shot action recognition methods are emerging and thriving. Existing few-shot action recognition methods mainly use pre-trained backbone models on image benchmarks (e.g., ImageNet-1k~\cite{ImageNet1k} and CLIP~\cite{CLIP_ICML}), which focus on the frame-level alignment between query and support videos in few-shot learning (FSL). Some early researches~\cite{2019TARN,2020otam,2021Few2} estimate temporal alignment for frame-level features to match the query videos and support set. TRX~\cite{2021Temporal} leverages an attentional mechanism to match each query video with all videos in the support set. HyRSM~\cite{2022Hybrid} introduces a hybrid relation module and designs a Bi-MHM for flexible matching. STRM~\cite{2021Spatio} proposes a spatio-temporal enrichment module to analyze spatial and temporal contexts. MASTAF~\cite{Liu_2023_WACV} uses self-attention and cross-attention modules to increase the inter-class variations while decrease the intra-class variations. MoLo~\cite{wang2023molo} learns long-range temporal context and motion cues for comprehensive few-shot matching. CLIP-FSAR~\cite{wang2023clip} devises a video-text contrastive objective and proposes a prototype modulation to fully utilize the rich semantic priors in CLIP. Different from the aforementioned works, our method aims to perform an effective yet efficient temporal-aware model tuning on the pre-trained frozen backbones to realize CDFSAR tasks.

\subsection{Cross-Domain Few-Shot Action Recognition}
Cross-domain few-shot learning requires base and test data from different domains. BS-CDFSL~\cite{2020A} first introduces an image benchmark for cross-domain few-shot learning, and early studies handle cross-domain action recognition by mainly focusing on deep feature learning and cross-domain learning~\cite{DBLP:journals/access/GaoHZZW18}. Meanwhile, as supplements, previous works~\cite{2020A2,DBLP:journals/tcyb/GaoZZCLC22} also introduce some source-target data pairs to evaluate the performance of CDFSAR. Recently, SEEN~\cite{WANG2023103737} proposes to integrate supervised learning with an auxiliary self-supervised contrastive learning to tackle the issue of domain gap lying in CDFSAR task. For above works~\cite{2020A2,DBLP:journals/tcyb/GaoZZCLC22,WANG2023103737}, there exist some shared classes lying in the constructed source-target data pairs, which however, is not expected in CDFSL task. CDFSL-V~\cite{10378593} proposes a new benchmark to solve this problem by removing all overlapping classes between the source and target datasets. DMSD~\cite{guo2024dmsd} designs two branches called the original-source and the mixed-source branches for meta-training based on the pipeline of CDFSL-V. But different from pair-wise joint training studied in previous CDFSAR methods~\cite{WANG2023103737,10378593,guo2024dmsd}, our proposed TAMT develops a decoupled paradigm to avoid frequent retraining in case of one source data and multiple target data, while proposing an HTTN method to effectively and efficiently adapt pre-training models for the target domain. Experimental comparisons (Sec.~\ref{sec:sota}) show our TAMT significantly outperforms existing counterparts.

\section{Method}
In this section, we first provide a brief definition of CDFSAR task. Then, we show the overview of our decoupled TAMT paradigm, which pre-trains models on source data while fine-tuning the pre-trained models on target one. Finally, we detailedly introduce the proposed hierarchical temporal tuning network (HTTN) for model tuning on target domain, which consists of local Temporal-Aware Adapters (TAA) and Global Temporal-aware Moment Tuning (GTMT). 
\subsection{Problem Formulation} 
CDFSAR task aims to develop an FSAR model for mitigating the side effect brought by domain gap between the source dataset $\mathcal{S}_{CD}$ and the target dataset $\mathcal{T}_{CD}$. In the context of cross-domain, the model could be trained on well-annotated $\mathcal{S}_{CD}$ and $\mathcal{T}_{CD}$ with few annotated samples. After that, the transferring performance of the proposed method is evaluated on target domain $\mathcal{T}_{CD}$.
\begin{figure*}[tb]
  \centering
  \begin{subfigure}{0.44\linewidth}
  \centering
    \includegraphics[height=5.255cm]{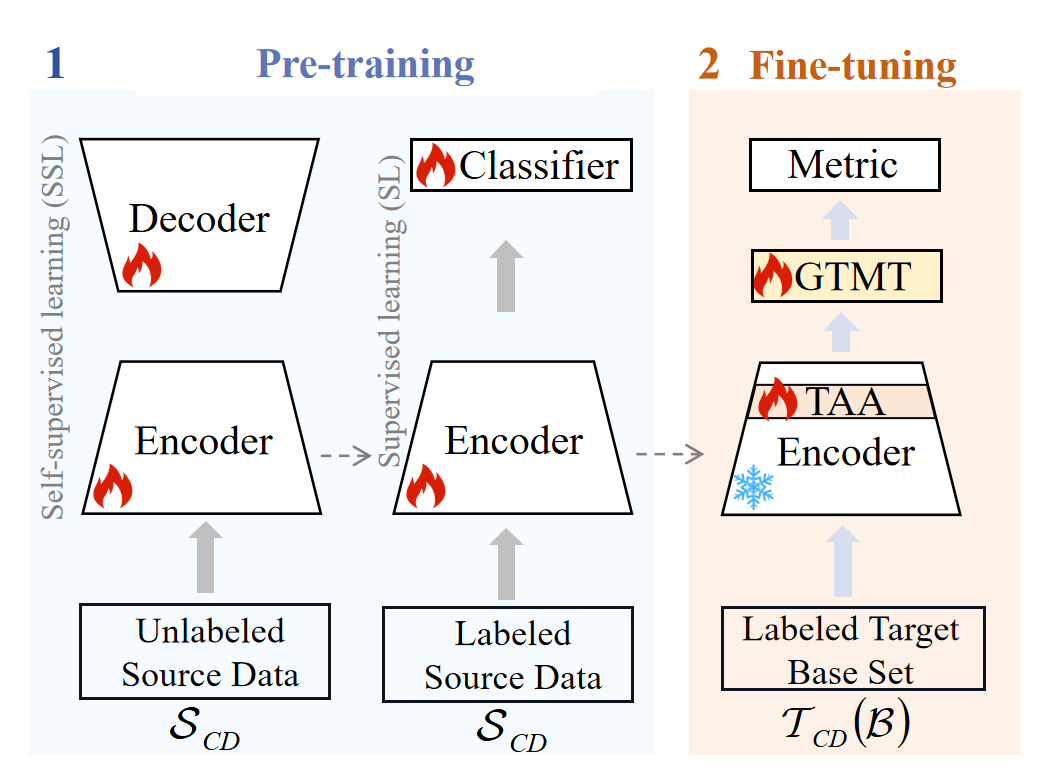}
    \caption{Overview of TAMT}
    \label{fig2:a}
  \end{subfigure}
  \hfill
  \begin{subfigure}{0.52\linewidth}
  \centering
    \includegraphics[height=5.255cm]{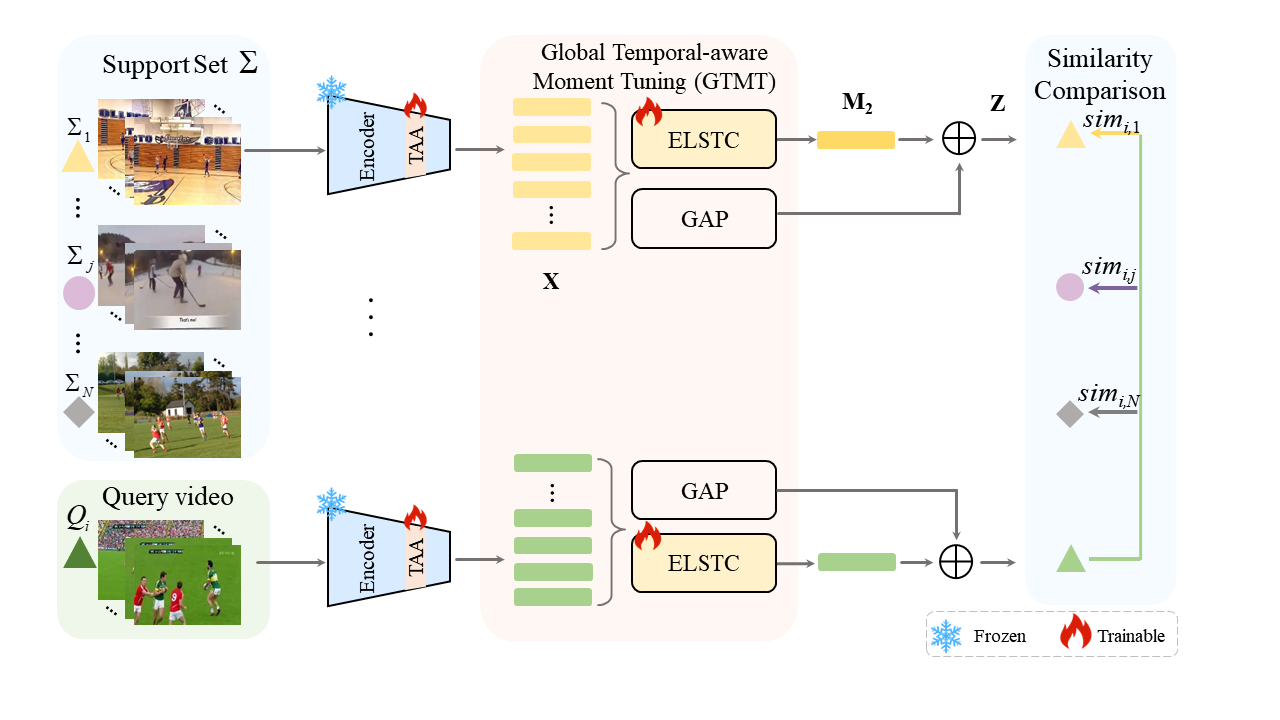}
    \caption{HTTN for Few-Shot Adaptation}
    \label{fig2:b}
  \end{subfigure}
  \caption{(a) Overview of our TAMT paradigm, which pre-trains the models on $\mathcal{S}_{CD}$ and fine-tunes them on $\mathcal{T}_{CD}$. Specifically, for pre-training stage, the model is first optimized with a reconstruction-based SSL solution, while the encoder $\mathcal{E}$ is post-trained with the SL objective. Subsequently, the pre-trained $\mathcal{E}$ is fine-tuned for few-shot adaptation on $\mathcal{T}_{CD}$ by using our  HTTN. (b) HTTN for few-shot adaptation, where a metric-based is used for few-shot adaptation. Particularly, our HTTN consists of local Temporal-Aware Adapters (TAA) and Global Temporal-aware Moment Tuning (GTMT).}
  \label{fig:TAMT}
\end{figure*}
In the target-domain $\mathcal{T}_{CD}$, the pre-trained model is evaluated on its novel (test) set $\mathcal{N}$ under FSL protocol, by providing training samples from its base (training) set $\mathcal{B}$, w.r.t., $\mathcal{T}_{CD} = \mathcal{B} \bigcup \mathcal{N}$.
For one FSL inference unit (dubbed as task or episode), it consists of unknown query videos $\{\mathcal{Q}_1, \cdots, \mathcal{Q}_{U}\}$, and an annotated support set $\Sigma$. For $N$-way $K$-shot setting, each episode involves $N$ categories and each category has $K$ samples in $\Sigma$. The final goal of CDFSAR is to accurately classify each query video $\mathcal{Q}_i$ by leveraging the limited data available in the support set $\Sigma$.
Particularly, to assess the transferring performance in a convincing way, the classes are non-overlapping in $\mathcal{S}_{CD}$ and $\mathcal{T}_{CD}$, i.e., $\mathcal{S}_{CD} \bigcap \mathcal{T}_{CD}  = \varnothing$, and $\mathcal{N} \bigcap \mathcal{B}  = \varnothing$ for FSL.

\subsection{Overview of Temporal-aware Model Tuning}\label{3.2}

Compared to FSAR, CDFSAR is further challenged by domain gap lying in source-to-target transfer learning. Previous works~\cite{WANG2023103737,10378593} develop some joint training paradigms on source and target data to mitigate side effect of domain gap. As shown in Fig.~\ref{fig:pipeline}, joint training paradigm generally suffers from model retraining in case of one source and multiple target data. Besides, they take no full advantage of pre-trained models by using some straightforward few-shot adaptation methods, potentially limiting recognition performance. To solve above issues, we propose a decoupled training paradigm, namely TAMT. As shown in Fig.~\ref{fig:TAMT} (a), our TAMT can be summarized as two phases: pre-training on $\mathcal{S}_{CD}$ and fine-tuning on $\mathcal{T}_{CD}$. Specifically, the model is first pre-trained on $\mathcal{S}_{CD}$ to learn the knowledge from source domain. Subsequently, it is fine-tuned on $\mathcal{T}_{CD}$ to perform transfer learning on target domain. The details are as follows.


\noindent \textbf{Pre-training on Source Data.} In this work, our TAMT pre-trains the models only on source data. To consider both generalization and representation abilities of pre-trained model, our TAMT develops a two-stage pre-training strategy. Inspired by success of self-supervised learning (SSL) on video pre-trained  models~\cite{2022VideoMAE}, we first introduce the reconstruction-based SSL solution to train our models for capturing general spatio-temporal structures lying videos, helping our pre-trained models can be well generalized to various downstream tasks. However, such SSL solution usually focuses on the fundamental features~\cite{he2022masked}, while neglecting high-level semantic information~\cite{videomaev2, iBoT_ICLR21, DINOV2,qi2023recon}, and limiting representation or discriminative abilities of pre-trained models. Existing works~\cite{iBoT_ICLR21, DINOV2,qi2023recon} make attempts to combine reconstruction-based SSL with self-supervised contrastive learning to improve discriminative ability of pre-trained models. By considering the samples are well annotated on source data, we simply incorporate a supervised learning (SL) after SSL to enhance the representation ability of pre-trained models. To be specific, the encoder $\mathcal{E}$ of the model is first trained with reconstruction-based SSL, and then it is optimized with recognition objectives on annotated $\mathcal{S}_{CD}$. As such, our two-stage pre-training strategy potentially achieves generalization and representation trade-off, where both SSL and SL play key roles in the final performance. More analysis can refer to Sec.~\ref{sec:ablation}.

\noindent \textbf{Fine-tuning on Target Data.} By considering the issue of domain shift between source data and target one, we propose a hierarchical temporal tuning network (HTTN), aiming to effectively and efficiently adopt the pre-trained model $\mathcal{E}$ to target domain. In particular, we construct our HTTN by using a metric-based FSL pipeline~\cite{snell2017prototypical}. To fully explore the potential of the frozen pre-trained models in transferring to target domain, we present local temporal-aware adapters (TAA) and a global temporal-aware moment tuning (GTMT) to recalibrate the intermediate features and generate powerful video representations according to few annotated samples on target domain, respectively. The details of our HTTN will be described in the following subsection. 

\subsection{Hierarchical Temporal Tuning Network}\label{3.3}
To perform few-shot adaptation of pre-trained models on target domain $\mathcal{T}_{CD}$, we propose a Hierarchical Temporal Tuning Network (HTTN). As depicted in Fig.~\ref{fig:TAMT} (b), our HTTN integrates several \textit{local} Temporal-Aware Adapters (TAA) into last-$L$ transformer blocks of pre-trained model $\mathcal{E}$, and insert a Global Temporal-aware Moment Tuning (GTMT) module with efficient long-short temporal covariance (ELSTC) at the end of $\mathcal{E}$. Given an input video, the features are extracted by the frozen $\mathcal{E}$, which are recalibrated by our TAA modules and subsequently fed into GTMT to generate final representation. Ultimately, the representations derived from query and support videos are compared using a similarity metric, which serves as logits for training and inference. 

\begin{figure*}[tb]
  \centering
  \begin{subfigure}{0.64\linewidth}
  \centering
    \includegraphics[height=6.2cm]{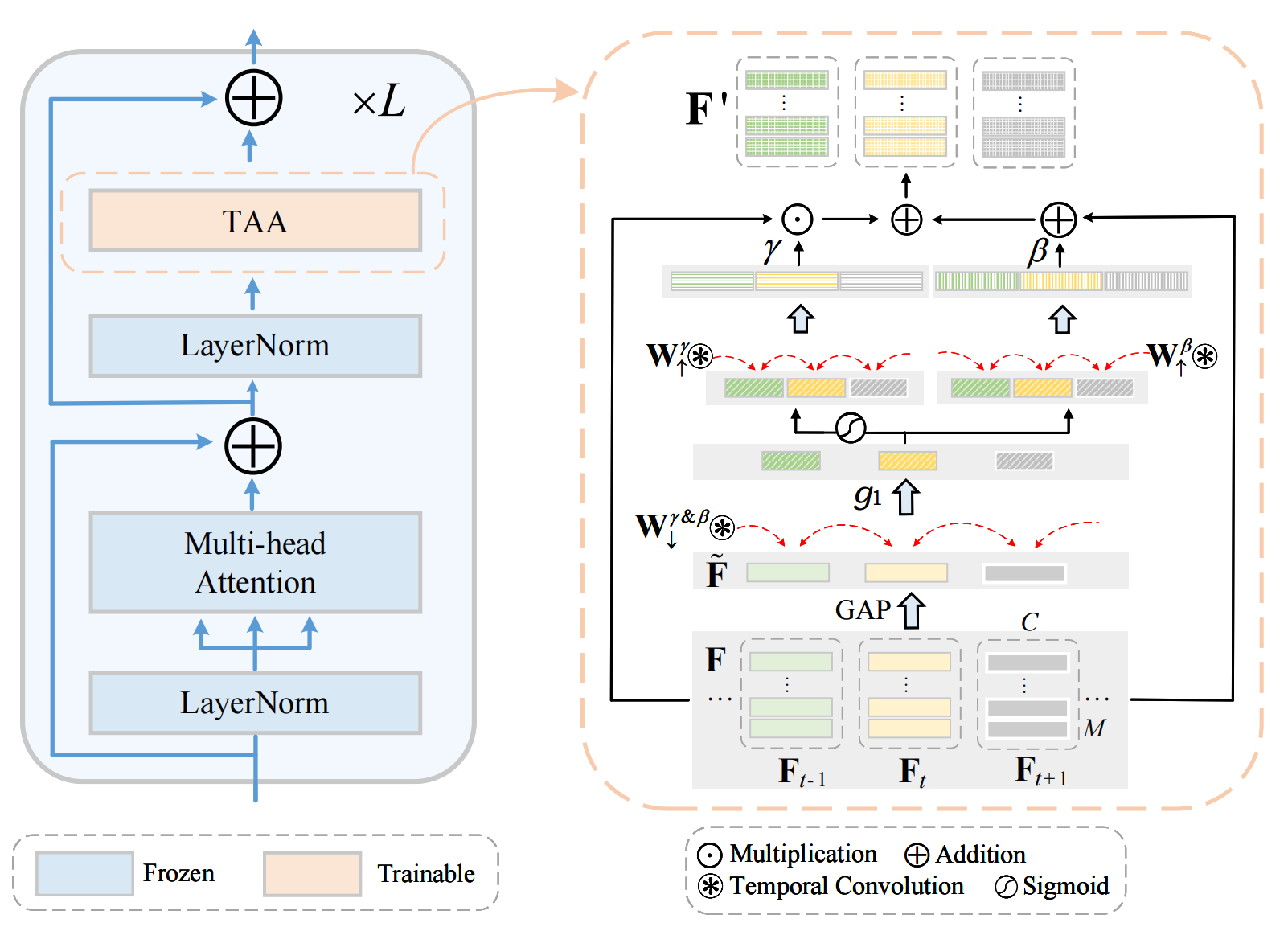}
    \caption{TAA}
    \label{fig:TAA}
  \end{subfigure}
  \hspace{1pt}
  \begin{subfigure}{0.32\linewidth}
  \centering
    \includegraphics[height=6.2cm]{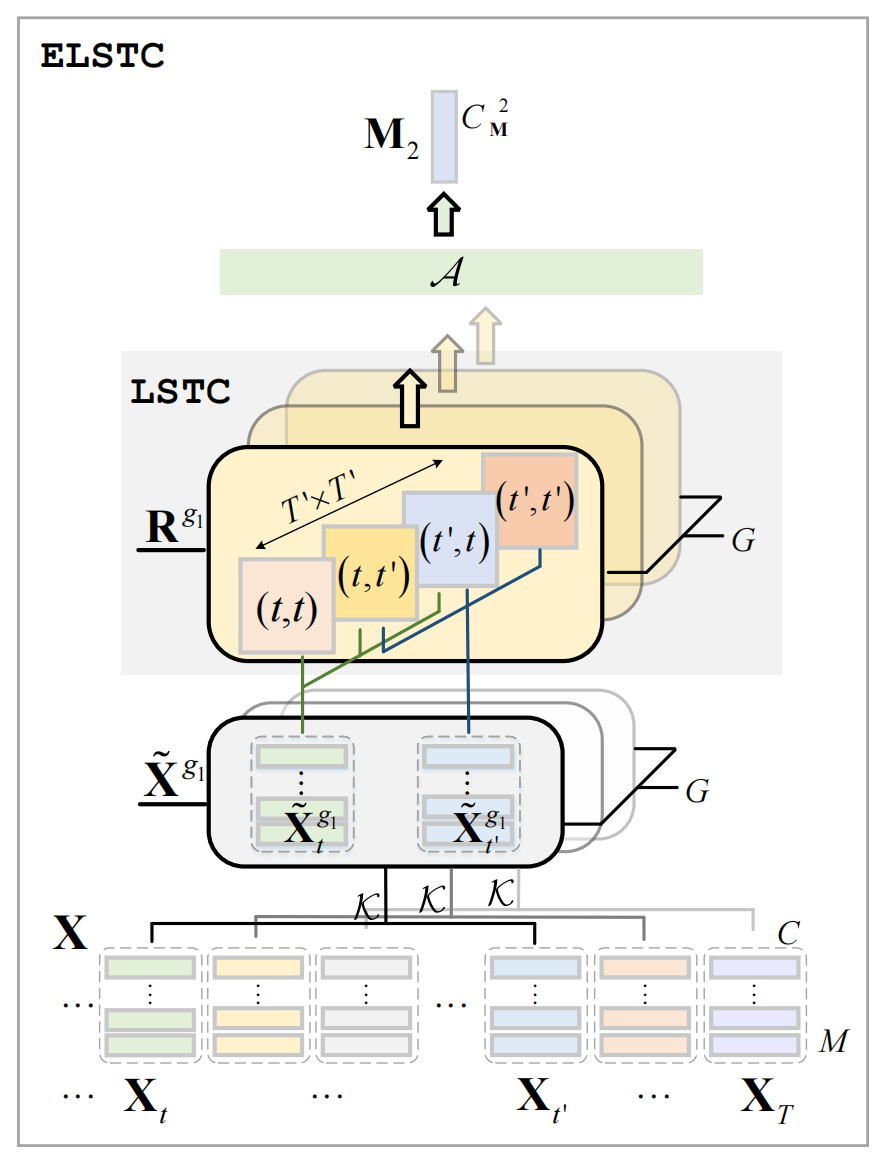}
    \caption{ELSTC}
    \label{fig:MHTC2P}
  \end{subfigure}
  \caption{Overview of our proposed Hierarchical Temporal Tuning Network (HTTN), where (a) local temporal-aware adapters (TAA) are inserted into the last $L$ transformer blocks to recalibrate the intermediate features of frozen pre-training models in an efficient manner. At the end of HTTN, a Global Temporal-aware Moment Tuning (GTMT) module with efficient long-short temporal covariance (ELSTC) is used to obtain powerful video representations for improving matching performance.}
  \label{fig:HTT}
\end{figure*}
\noindent \textbf{Local Temporal-Aware Adapter (TAA).}
In the decoupled training protocol of TAMT, it is important to utilize target domain for tuning source pre-trained model. However, the conventional full fine-tuning (FFT) strategy will optimize all of the parameters, bringing high training consumption and potentially posing the risk of overfitting, particularly in the few-shot learning scenario. As suggested in previous works for NLP and image classification tasks~\cite{houlsby2019parameter, Lian_2022_SSF}, we introduce a parameter-efficient approach for recalibrating video features in a temporal-aware manner.

Given a certain intermediate-layer feature $\mathbf{F}\in\mathbb{R}^{T\times M\times C}$, where $T, M, C$ denotes temporal length, token number in one frame and channel number, respectively. TAA introduces a few learnable scale and bias parameters for features of each frame, which can be written as follows:
\begin{equation}\label{eq:taa}
\mathbf{F}' = \gamma \odot \mathbf{F} \oplus \beta,
\end{equation}
where $\odot$ and $\oplus$ represent the multiplication and addition operations, respectively. For the sake of convenience, here we omit the expansion operation along $M$ dimension for scale $\gamma\in\mathbb{R}^{T\times C}$ and bias $\beta\in\mathbb{R}^{T\times C}$.
Particularly, $\gamma$ and $\beta$ indicate the temporal cues of $\mathbf{F}$ decided by transformation functions $\mathcal{W}$ and $\mathcal{G}$ respectively. Therefore, we have
\begin{equation}\label{eq:gamma}
\gamma=\mathcal{W}\left(\widehat{\mathbf{F}}\right)=\mathbf{W}_{\uparrow}^{\gamma}\circledast g_1\left(\mathbf{W}_{\downarrow}^{\gamma} \circledast \widehat{\mathbf{F}}\right),
\end{equation}
\begin{equation}\label{eq:3}
\beta=\mathcal{G}\left(\widehat{\mathbf{F}}\right)=\mathbf{W}_{\uparrow}^{\beta}\circledast g_2\left(\mathbf{W}_{\downarrow}^{\beta} \circledast \widehat{\mathbf{F}}\right),
\end{equation}
where $\widehat{\mathbf{F}}\in \mathbb{R}^{T\times C}$ presents the output of global average pooling on $\mathbf{F}$. By taking Eqn.~(\ref{eq:gamma}) as an example, function $\mathcal{W}$ is efficiently implemented by a two-layer temporal convolution with temporal kernel $k_t>1$ (denoted as $\circledast$), effectively capturing the temporal information. Particularly, for parameter efficiency, $\mathcal{W}$ is realized by involving a dimensionality reduction operation $C\rightarrow \frac C {\rho} \rightarrow C$ with a hyper parameter of $\rho$ and parameters of $\mathbf{W}^{\gamma}_{\downarrow}$ and $\mathbf{W}^{\gamma}_{\uparrow}$. And the operation $g_{*}$ is on behalf of the activation function, aiming to enhance the non-linear relation, among which $g_1$ is a ReLU layer followed by a sigmoid function, $g_2$ is a single ReLU layer.  
Moreover, for further parameter efficiency, $\mathbf{W}^{\gamma}_{\downarrow}$ and $\mathbf{W}^{\beta}_{\downarrow}$ are implemented with a parameter-shared $\mathbf{W}^{\gamma \&\beta}_{\downarrow}$, w.r.t., $\mathbf{W}^{\gamma \&\beta}_{\downarrow}:=\mathbf{W}^{\gamma}_{\downarrow}:=\mathbf{W}^{\beta}_{\downarrow}$. The structure of TAA is illustrated in Fig.~\ref{fig:HTT} (a). Practically, $\rho$ is set to 4 and $k_t=3$ throughout all of experiments in this work.

Note that, as shown in Fig.~\ref{fig:HTT} (a), our light-weight TAA is generally embedded into last-$L$ transformer blocks in a plug-and-play manner. Thus, HTTN can be efficiently tuned by freezing most of the pre-trained parameters, and only partially learning a few parameters, which provides a both parameter- and memory-efficient tuning solution for the pre-trained model. Simple adapters~\cite{houlsby2019parameter, Lian_2022_SSF} focus only on modeling spatial information, our TAA additionally learns temporal information. And different from other temporal adapters ~\cite{chen2023tem, pan2022st}, which respectively uses an autoregressive task and 3D depth-wise convolution for temporal alignment and adapter, our TAA efficiently re-scales and translates video features in a temporal-aware way.

\noindent \textbf{Global Temporal-Aware Moment Tuning (GTMT).}
In general, the FSL task can be regarded as a comparison problem between query and support representations $\mathbf{Z}_{\mathcal{Q}_{i}}$ and $\mathbf{Z}_{\mathcal{S}_{j}}$ in query video set $\mathcal{Q}_i$ and support video set $\mathcal{S}_j$, i.e., $sim_{i,j}=\mathcal{D}\left(\mathbf{Z}_{\mathcal{Q}_{i}},\mathbf{Z}_{\mathcal{S}_{j}}\right)$ where $\mathcal{D}$ is a pre-defined metric. 
Intuitively, a powerful representation $\mathbf{Z}$ will help the matching performance. By considering that previous works~\cite{WANG2023103737,10378593} take no merit of rich statistical information inherent in deep features, our HTTN proposes Global Temporal-Aware Moment Tuning (GTMT) method to exploit probability distribution for modeling video features, which can effectively characterize statistics of features and provide powerful representations~\cite{gao2023tuning, bishop2006pattern}.
Let $\mathbf{X}\in\mathbb{R}^{T\times M \times C}$ be the features from the last transformer block of HTTN, the probability distribution of features $\mathbf{X}$ can be approximately portrayed by feature moment~\cite{gao2023tuning}. Let $\mathbf{\Phi}_\textbf{X}(u)$ be characteristic function of features $\mathbf{X}$ with argument $u\in\mathbb{R}$, $\mathbf{Z}$ can be written as:
\begin{equation}\label{eq:taylor}
\mathbf{Z}:=\mathbf{\Phi}_{\mathbf{X}}(u)=1+\alpha_1 \mathbf{M}_1+\alpha_2 \mathbf{M}_2+\cdots=1+\sum_{p=1}^{\infty} \alpha_p \mathbf{M}_p,
\end{equation}
where $\mathbf{M}_p$ indicates the $p^{th}$-order moment of $\textbf{X}$ and $\alpha_p\in\mathbb{R}$ is the coefficient. For the consideration of computational cost, $p$ is maximum to 2, which involves the zero-, first-, and second-order moments. 

By considering the temporal dynamic in the video features, we compute the first- and second-order moment $\mathbf{M}_1$ and $\mathbf{M}_2$ in a temporal manner. To be specific, $\mathbf{M}_1$ can be obtained by the global average pooling (GAP) layer:
\begin{equation}\label{eq:GAP}
\mathbf{M}_1=\mathtt{GAP}\left(\mathbf{X}\right)=\frac 1 {TM} \sum_{t=1}^{T} \sum_{m=1}^{M} \mathbf{X}_{t,m}.
\end{equation}
However, a straightforward implementation of the second-order moment $\mathbf{M}_2$, referring to a simple temporal covariance with concurrently considering interaction across all frames (termed as TCov), typically needs $\left(TC\right)^2$-dimensional computation consumption.
Thereby, we propose an Efficient Long-Short Temporal Covariance layer (ELSTC), 
\begin{equation}\label{eq:MHTC2P}
\mathbf{M}_2=\mathtt{ELSTC}\left(\mathbf{X}\right)=\mathcal{A}\left(\mathtt{LSTC}\left(\mathbf{X}^{g_1}\right);
\cdots;\mathtt{LSTC}\left(\mathbf{X}^{g_G}\right)\right).
\end{equation}
For high efficiency, the sequence feature $\mathbf{X}$ is split into $G$ groups along temporal dimension and can be rewritten as: $\mathbf{X}$ = $\left[\mathbf{X}^{g_1};\cdots;\mathbf{X}^{g_G}\right]$, where notation $[\cdot]$ indicates concatenation. For group $g_e$, $\mathbf{X}^{g_e}\in\mathbb{R}^{T'\times M\times C}$ is with temporal length $T'=\frac T G$, leading to reduce the computation consumption $G$ times. In each group $g_e$, aiming for temporal-aware modeling, we devise LSTC to compute long-short temporal covariance as follows:
\begin{equation}
\mathtt{LSTC}\left(\mathbf{X}^{g_e}\right)= \left\{\mathbf{R}_{t,t'}^{g_e}\right\}_{{t,t'}}^{T'^2};
\end{equation}
\begin{equation}
\mathbf{R}_{t,t'}^{g_e}=\frac{1}{M} \sum_{m=1}^{M}\widetilde{\mathbf{X}}_{t,m}^{{g_e}^{\top}}\widetilde{\mathbf{X}}_{t',m}^{g_e}; 
\widetilde{\mathbf{X}}=\mathcal{K}\left(\mathbf{X}\right).
\end{equation}
In group $g_e$, the covariance matrix $\mathbf{R}_{t,t'}^{g_e}\in\mathbb{R}^{\frac C \tau \times \frac C \tau }$ captures temporal correlation between $t$-th and $t'$-th frame of $\widetilde{\mathbf{X}}\in\mathbb{R}^{T'\times M \times  \frac C \tau}$. The feature $\widetilde{\mathbf{X}}$ is a transformation of $\mathbf{X}$ with an MLP layer $\mathcal{K}$, bringing the dimension reduction by a hyper parameter $\tau$: $C\rightarrow \frac C {\tau}$. $\top$ represents transposition operation. 

In particular, for one covariance matrix $\mathbf{R}_{t,t'}^{g_e}$ in group $g_e$, the timestamps $t$ and $t'$ always have a temporal gap $\Delta$, ranging from 0 to $\left(T-G\right)$ with an interval of G.
For $\Delta=0$, $\mathbf{R}_{t,t}^{g_e}$ indicates the static appearance information of $\widetilde{\mathbf{X}}_{t}^{g_e}$, and for other $\Delta\neq 0$, $\mathbf{R}_{t,t'}^{g_e}$ outputs the temporal cross-covariance of $\widetilde{\mathbf{X}}_{t}^{g_e}$ and $\widetilde{\mathbf{X}}_{t+\Delta}^{g_e}$. As a result, the output of $\mathtt{LSTC}\left(\mathbf{X}^{g_e}\right)$ describes the various temporal correlations from \textit{short-term} (one frame) static appearance to \textit{long-term} crossing $\left(T-G\right)$ frames motion information.

Furthermore, the outputs derived from $G$ groups are ultimately summarized with $\mathcal{A}$, generating a holistic video representation $\mathbf{M}_2$.
To the sake of clarity, the output of LSTC for group $g_e$ is rewritten as $\mathbf{Y}^{g_e}\in\mathbb{R}^{{T'\frac C {\tau}}\times T'\frac C {\tau} \times 1}$ by concatenating all $\{\mathbf{R}_{*,*}^{g_e}\}$ in group $g_e$. 
And then, $\mathbf{M}_2$ is:
\begin{equation}
\mathbf{M}_2 = \mathcal{A}\left(\mathbf{Y}^{g_1},\cdots,\mathbf{Y}^{g_G}\right),
\end{equation}
where the indication $\mathcal{A}$ denotes two convolutional layers with $k_c\times k_c$ kernel, with each followed by a BN layer and ReLU activation function. By setting the proper stride and output channel, the dimension of $G$-group output is changed from ${T'\frac C {\tau}}\times T'\frac C {\tau}\times G$ to $C_{\mathbf{M}}\times C_{\mathbf{M}}\times 1$, and vectorized to $C_{\mathbf{M}}^{2}\times 1$ ultimately. By omitting the constant zero-order in Eqn.~(\ref{eq:taylor}), the final representation of our HTTN is expressed by combining the first and second-order moment as follows:
\begin{equation}
\mathbf{Z}=\mathcal{H}\left(\mathbf{M}_2\right) \oplus \mathbf{M}_1,
\end{equation}
where a linear projection $\mathcal{H}$ is used to align the dimension of $\mathbf{M}_2$ with that of $\mathbf{M}_1$, i.e., $C_{\mathbf{M}}^2\rightarrow C$. In this work, $k_c=3, \tau=6, C_{\mathbf{M}}=64$ for all experiments.
\section{Experiments}
We extensively compare our TAMT with state-of-the-arts on both CDFSAR and FSAR tasks (see supplementary material), and conduct the ablation study on CDFSAR task. 
\subsection{Experimental Settings}
\textbf{Datasets.} For CDFSAR task, we use the K-400~\cite{2017The}, K-100~\cite{2018Compound}, SSV2~\cite{2017The2}, Diving~\cite{2018RESOUND} and UCF~\cite{2012UCF101} as the source domains $\mathcal{S}_{CD}$, which transfer to following five target domains $\mathcal{T}_{CD}$: HMDB~\cite{2011HMDB}, SSV2, Diving, UCF and RareAct~\cite{2020RareAct}. 
For the source datasets, we follow the non-overlapping setting protocol~\cite{10378593} between $\mathcal{S}_{CD}$ and $\mathcal{T}_{CD}$ in cross-domain scenario. Specifically, source datasets K-400 and K-100 are removed some shared classes with UCF and HMDB, resulting in 364 and 61 categories retained respectively. 
For the target datasets, we utilize established splits for HMDB, SSV2, Diving and UCF as outlined in previous studies~\cite{10378593,Liu_2023_WACV,2021Learning,2022Hybrid,wang2023molo,KP_FSL}. For RareAct database, we split the base, validation and novel set with 48, 8 and 8 classes. For FSAR task, TAMT is evaluated on SSV2, HMDB and UCF, whose splits follow their configurations in CDFSAR.

\begin{table*}[t]
    \centering
    \fontsize{8}{6.5}\selectfont
    \setlength{\tabcolsep}{13.75pt}
    \renewcommand{\arraystretch}{1.392}
    \scalebox{1.07}{
    \begin{tabular}{l|c|c|c|c|c|c|c}\hline
    \multirow{2}*{Method} & \multirow{2}*{Source} & \multicolumn{6}{c}{Target} \\
    \cline{3-8}
       &   & HMDB & SSV2 & Diving & UCF & RareAct & Average \\ \hline
      \rowcolor{white} 
      \textcolor{gray}{STARTUP++~\cite{phoo2020self}} &  & \textcolor{gray}{44.71} & \textcolor{gray}{39.60} & \textcolor{gray}{14.92} & \textcolor{gray}{60.82} & \textcolor{gray}{45.22} & \textcolor{gray}{41.05} \\
      \rowcolor{white} \textcolor{gray}{DD++~\cite{islam2021dynamic}} & & \textcolor{gray}{48.04} & \textcolor{gray}{44.50} & \textcolor{gray}{16.23} & \textcolor{gray}{63.26} & \textcolor{gray}{47.01} & \textcolor{gray}{43.81} \\
      \rowcolor{white} \textcolor{gray}{STRM~\cite{2021Spatio}} & & \textcolor{gray}{24.98} & \textcolor{gray}{35.01} & \textcolor{gray}{16.69} & \textcolor{gray}{42.33} & \textcolor{gray}{39.01} & \textcolor{gray}{31.60} \\
      \rowcolor{white} \textcolor{gray}{HYRSM~\cite{2022Hybrid}} &K-400 & \textcolor{gray}{29.81} & \textcolor{gray}{40.09} & \textcolor{gray}{17.57} & \textcolor{gray}{45.65} & \textcolor{gray}{44.27} & \textcolor{gray}{35.49} \\

      CDFSL-V~\cite{10378593} & & 53.23  & 49.92  & 17.84 & 65.42 & 49.80 & 47.24\\
      TAMT~(Ours) & & $\textcolor{blue}{\textbf{74.14}}$ & $\textcolor{blue}{\textbf{59.18}}$ & $\textcolor{blue}{\textbf{45.18}}$ & $\textcolor{blue}{\textbf{95.92}}$ & $\textcolor{blue}{\textbf{67.44}}$ & $\textcolor{blue}{\textbf{68.37\textsubscript{(+21.13)}}}$\\
      TAMT*~(Ours) & &  \textbf{\textcolor{red}{77.82}} & \textbf{\textcolor{red}{64.20}} & \textbf{\textcolor{red}{49.16}} & \textbf{\textcolor{red}{97.08}} & \textbf{\textcolor{red}{73.31}} & \textbf{\textcolor{red}{72.31}}\\ \hline

      \textcolor{gray}{STARTUP++~\cite{phoo2020self}} & \multirow{7}*{K-100} & \textcolor{gray}{24.97}  & \textcolor{gray}{15.16} & \textcolor{gray}{14.55} & \textcolor{gray}{32.20} & \textcolor{gray}{31.77} & \textcolor{gray}{23.73}\\
      \textcolor{gray}{DD++~\cite{islam2021dynamic}} &   & \textcolor{gray}{25.99} & \textcolor{gray}{16.00} & \textcolor{gray}{16.24} & \textcolor{gray}{34.10} & \textcolor{gray}{31.20} & \textcolor{gray}{24.71}\\
      SEEN*\dag~\cite{WANG2023103737} &  & 52.80  &  31.20  & 40.90 &  79.60 & 50.20 & 50.94\\
      CDFSL-V~\cite{10378593} & & 29.80  & 17.21  & 16.37 & 36.53 & 33.91 & 26.76\\
      DMSD*\dag~\cite{guo2024dmsd} & & 54.90  & 32.10  & \textcolor{blue}{\textbf{42.28}} & 81.90 & \textcolor{blue}{\textbf{53.30}} & 52.90\\
      TAMT~(Ours) & & $\textcolor{blue}{\textbf{61.76}}$ & $\textcolor{blue}{\textbf{48.90}}$ & 38.33 & $\textcolor{blue}{\textbf{87.76}}$ & 52.81 & $\textcolor{blue}{\textbf{57.91\textsubscript{(+31.15)}}}$\\
      TAMT*~(Ours) & & \textbf{\textcolor{red}{70.87}} & \textbf{\textcolor{red}{58.38}} & \textbf{\textcolor{red}{44.15}} & \textbf{\textcolor{red}{93.69}} & \textbf{\textcolor{red}{61.67}} & \textbf{\textcolor{red}{65.75}}\\ \hline

      CDFSL-V~\cite{10378593} & \multirow{2}*{SSV2} & 29.86  &  - & 17.60 & 33.30 & 35.25 & 29.00\\
      TAMT~(Ours) & & $\textcolor{red}{\textbf{63.66}}$ & -  & $\textcolor{red}{\textbf{38.75}}$ & \textcolor{red}{\textbf{83.45}} & $\textcolor{red}{\textbf{42.23}}$ & $\textcolor{red}{\textbf{57.02\textsubscript{(+28.02)}}}$\\ \hline

      CDFSL-V~\cite{10378593} & \multirow{2}*{Diving } & 28.45  & 17.46 & - &  31.98 & 34.11 & 28.00\\
      TAMT~(Ours) & & $\textcolor{red}{\textbf{45.18}}$ &  \textcolor{red}{\textbf{38.09}}  & - &  \textcolor{red}{\textbf{63.52}} & $\textcolor{red}{\textbf{36.88}}$ & $\textcolor{red}{\textbf{45.92\textsubscript{(+17.92)}}}$\\\hline

      \textcolor{gray}{STARTUP++~\cite{phoo2020self}} & \multirow{4}*{UCF} & \textcolor{gray}{23.56}  & \textcolor{gray}{-}  & \textcolor{gray}{14.84} &  \textcolor{gray}{-} & \textcolor{gray}{31.31} & \textcolor{gray}{23.24}\\
      \textcolor{gray}{DD++~\cite{islam2021dynamic}} &  &\textcolor{gray}{24.06}  & \textcolor{gray}{-} & \textcolor{gray}{16.15} &  \textcolor{gray}{-} & \textcolor{gray}{32.00} & \textcolor{gray}{24.07}\\
      CDFSL-V~\cite{10378593} &  &\textcolor{blue}{\textbf{28.86}}  &  - & \textcolor{blue}{\textbf{16.07}} &  - & \textcolor{blue}{\textbf{33.91}} & \textcolor{blue}{\textbf{26.82}}\\
      TAMT~(Ours) & & $\textcolor{red}{\textbf{45.34}}$ &  -  & $\textcolor{red}{\textbf{33.38}}$ &  - & $\textcolor{red}{\textbf{41.08}}$ & $\textcolor{red}{\textbf{39.93\textsubscript{(+13.11)}}}$\\
   \hline
        \end{tabular} 
        }
    \caption{Comparison(\%) of state-of-the-arts on CDFSAR setting in terms of 5-way 5-shot accuracy, where five datasets (K-400, K-100, SSV2, Diving and UCF) are used as source data for transferring to five target datasets. All results are conducted with $112\times 112$ resolution by using ViT-S backbone, except methods marked by * ($224\times 224$ resolution) and marked by $\dag$ (backbone of ResNet-18).}
    \label{table1}
  \end{table*}

\noindent \textbf{Implementation Details.} We adopt VideoMAE~\cite{2022VideoMAE} as the backbone, which is respectively built on ViT-S or ViT-B architectures for CDFSAR and FSAR for fair comparison. 
If not specified otherwise, the input resolution is $112\times 112$ for ViT-S in CDFSAR and $224\times 224$ in ViT-B for FSAR. 
The video inputs of the model are set to 16 frames, and then they are reduced to 8 in the patch embedding stage before the first transformer block.
For optimization, we use SGD as the optimizer and adopt a cosine decay strategy to schedule the learning rate. The training epochs are set to 400, 140 and 40 for the SSL, SL and fine-tuning, respectively. In the pre-training phase, we employ the mean squared error and cross-entropy (CE) losses for SSL and SL, respectively. For the fine-tuning phase, we use CE loss. Euclidean distance is served as the metric function $\mathcal{D}$. We report accuracy in 5-way 1-shot and 5-way 5-shot settings on a single view, averaging 10,000 episodes for inference. Source code is available at
\url{https://github.com/TJU-YDragonW/TAMT}.

\newcommand{\rvgzl}[1]{\textcolor{blue}{#1}}

\subsection{Comparison with State-of-the-Arts}\label{sec:sota} 
To fully evaluate our TAMT in the CDFSAR task, we conduct experiments with 5 source datasets and 5 target datasets, which form a nearly one-\textit{vs.}-one cross-domain setting. Besides, we compare with state-of-the-art CDFSAR methods, which to our best knowledge cover all published CDFSAR  works. The results of different methods in terms of 5-way 5-shot accuracy are reported in Tab.~\ref{table1}, where the best and second best results are highlighted in \textcolor{red}{\textbf{red}} and \textcolor{blue}{\textbf{blue}} font, respectively. From Tab.~\ref{table1} we can conclude that (1) our TAMT outperforms existing methods by 13\%$\sim$31\% across all settings, leading to large performance gains. (2) On two widely used source datasets (K-400 and K-100), TAMT outperforms CDFSL-V~\cite{10378593} by an average of 21.13\% and 31.15\%  across five target datasets, respectively. For the source datasets SSV2, Diving and UCF, our TAMT achieves improvements of 28.02\%, 17.92\% and 13.11\% over CDFSL-V on average across different target datasets. (3) TAMT achieves performance improvements of 18.07\%, 27.18\%, 3.25\%, 14.09\%, and 11.47\% over SEEN~\cite{WANG2023103737} across five target domains, while outperforming DMSD~\cite{guo2024dmsd} with 15.97\%, 26.28\%, 1.87\%, 11.79\% and 8.37\%. Furthermore, compared to CDFSL-V, our TAMT with decoupled training paradigm has nearly 5× less training computational cost\footnote{ TAMT consumes 19 GPU days compared to CDFSL-V’s 88 GPU days when training on the K-400 source dataset across five target datasets.}. These results above clearly demonstrate that our TAMT provides a promising baseline for the CDFSAR task in terms of both efficiency and effectiveness.

\begin{table*}[t]
  \centering
  \fontsize{8}{6.5}\selectfont
  \setlength{\tabcolsep}{7pt}
  \renewcommand{\arraystretch}{1.4}
    \centering
    \begin{tabular}{cc|ccc:ccc:ccc|ccc}
    \hline
    \multicolumn{2}{c|}{Pre-tr} & \multicolumn{12}{c}{Tuning} \\
    \hdashline
       \multirow{2}*{SSL} &  \multirow{2}*{SL} & \multicolumn{3}{c:}{SSV2} & \multicolumn{3}{c:}{Diving} &  \multicolumn{3}{c|}{UCF} &   \multicolumn{3}{c}{Average} \\
       & & Frozen & FFT & TAMT & Frozen & FFT & TAMT &Frozen & FFT & TAMT & Frozen & FFT & TAMT   \\ \hline
     \checkmark  &  & 29.27 & 48.99 &  47.21 & 22.10 & 35.13 & 33.59 &55.30 & 80.92 &77.59 &35.56 & 55.01 & 52.80 \\
     & \checkmark & 34.91& 41.39& 45.15 & 27.15& 33.27& 37.96 & 88.41 & 89.36 & 89.73 & 50.16 & 54.67 & 56.48 \\
      \checkmark & \checkmark & \psp{40.45}& 55.99 & \hlours{59.18} & \psp{28.09} & 42.85 & \hlours{45.18} & \psp{94.69} & 94.95 & \hlours{95.92}& \psp{54.41} & 64.30 & \hlours{66.76} \\ 
      \hline
  \end{tabular}
  \caption{Results (\%) of various pre-training (Pre-tr) schemes and tuning strategies in terms of 5-way 5-shot accuracy (ViT-S as backbone). Memory: Training memory cost of FFT and TAMT. Parameter: Training parameters of FFT and TAMT.}
  \label{tab:pretrain}
\end{table*}
\begin{table}[t]
\centering
\fontsize{8}{6.5}\selectfont
\setlength{\tabcolsep}{11pt}
\renewcommand{\arraystretch}{1.439}
\begin{tabular}{l|ccccccc}
\hline
 Method & Memory & Parameters & Training Time 
 \\ \hline
FFT & 17.5G & 29.9M & 10.6h \\
TAMT & 1.9G & 2.8M & 7.3h \\ \hline
\end{tabular}
\caption{Efficiency Comparison for FFT and TAMT.}
\label{tab:parameter}
\end{table}
\begin{figure}[tb]
  \centering
  \begin{subfigure}{0.5\linewidth}
  \centering
    \includegraphics[height=3cm]{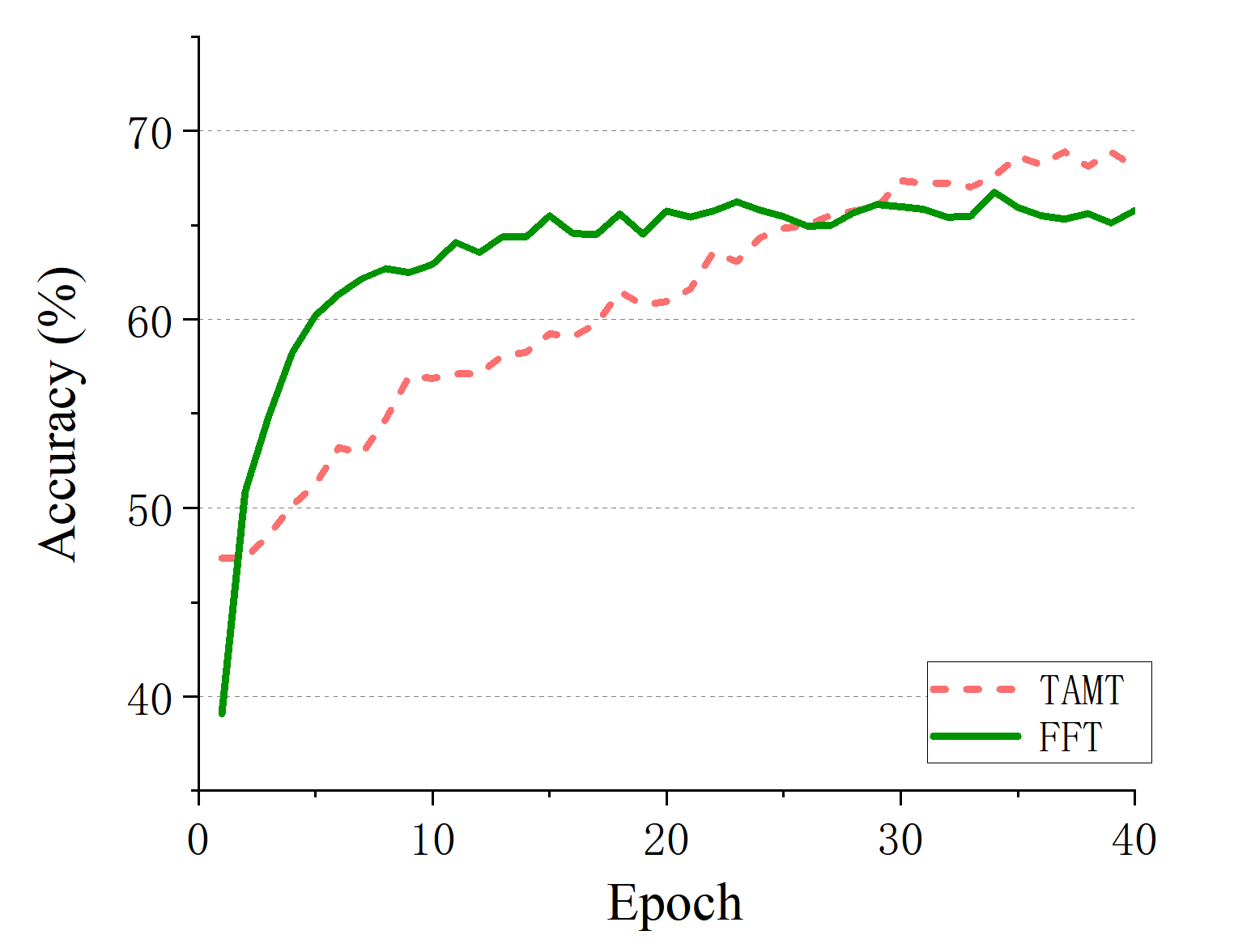}
    \caption{Training}
    \label{fig:TAA2}
    \end{subfigure}
  \begin{subfigure}{0.48\linewidth}
  \centering
  \includegraphics[height=3cm]{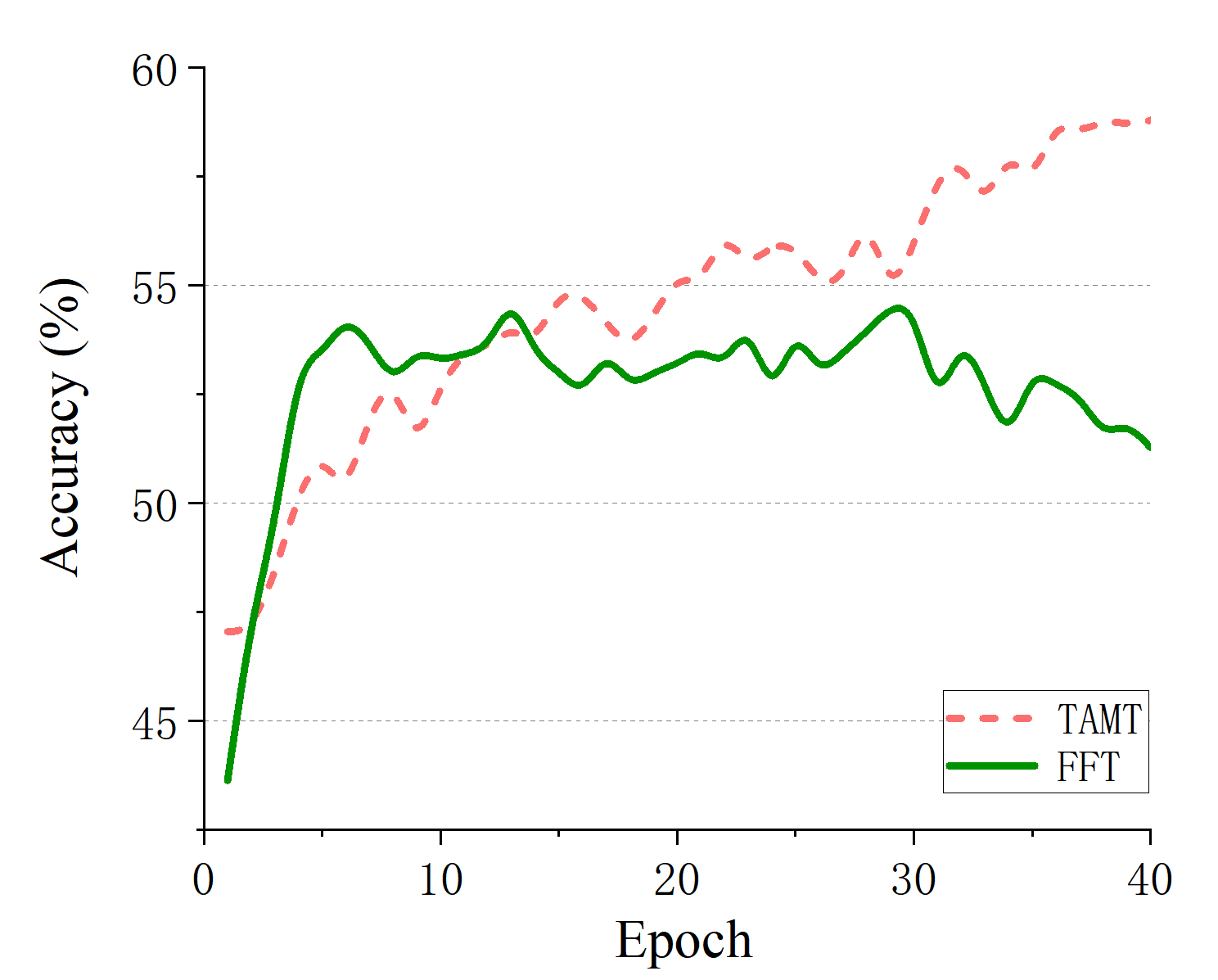}
  \caption{Validation}
  \label{fig:MHTC2P2}
\end{subfigure}
  \caption{Convergence curves of FFT and TAMT on SSV2 dataset.}
  \label{fig:curve}
\end{figure}
\subsection{Ablation Studies}\label{sec:ablation}
\begin{table*}[t]
    \centering
    \fontsize{8}{6.5}\selectfont
    \setlength{\tabcolsep}{6.5pt}
    \renewcommand{\arraystretch}{1.47}
    \begin{subtable}{0.5\linewidth}
    \centering
            \resizebox{0.97\textwidth}{!}{
                \begin{tabular}{cl|ccc|c}\hline
                    Adapter & Moment & SSV2 & Diving & UCF & Average\\ \hline
                    \multirow{2}{*}{None} & GAP  &  \psp{40.45} & \psp{28.09} & \psp{94.69} & \psp{54.41} \\
                    & GTMT & 53.41 & 42.87 & 94.97 & 63.58\\
                    \hdashline
                    \multirow{4}{*}{TAA}& GAP & 54.66 & 43.26  & 95.23 & 64.38 \\
                    & TCov & 56.72 & 43.68 & 95.16 & 65.19\\
                    & ELSTC & 58.56 & 43.90 & 95.37 & 65.94 \\
                     & GTMT & \hlours{59.18} &\hlours{45.18} & \hlours{95.92} & \hlours{66.76}\\
                    \hline
                \end{tabular}
                }
            \caption{Separated local or global modules.}
            \label{tab:component_sep}
    \end{subtable}
    \begin{subtable}{0.49\linewidth}
        \centering
            \resizebox{1\textwidth}{!}{
                \begin{tabular}{l|ccc|c}
                    \hline
                    Method & SSV2 & Diving & UCF & Average \\ 
                    \hline                    Adapter~\cite{houlsby2019parameter} &  52.01&40.72&95.05 & 62.59\\
                    SSF~\cite{Lian_2022_SSF} & 50.46&31.51&95.25 & 59.07 \\
                    TAM~\cite{liu2021tam} & 53.74 & 42.67 & 94.96 & 63.79 \\
                    ST-Adapter~\cite{pan2022st} & 54.26& 43.12 & 95.16 & 64.18 \\
                   TAMT (Ours) & \hlours{59.18} &\hlours{45.18} & \hlours{95.92} & \hlours{66.76}\\
                    \hline
                \end{tabular}
                }
                \caption{Adapter counterparts.}
            \label{tab:component_joinly}
       \end{subtable}
    \caption{Effect of core components (i.e., TAA \& GTMT) on HTTN, where the results (\%) in terms of 5-way 5-shot accuracy are reported.}
    \label{tab:HTTN_components}
\end{table*}

In this subsection, we conduct ablation studies to evaluate the effect of various components on three target datasets, i.e., SSV2, Diving and UCF, with K-400 serving as the source dataset. If not specified otherwise, the group number $G$ is 4, and TAA block number $L$ is 2 for the default option, whose ablation studies are reported in the supplemental material. The results highlighted with the same color indicate the identical deployment and experimental setting. To compare the training efficiency, we compare the GPU memory allocation required for training one episode (with 85 instances) on a server equipped with single NVIDIA TiTAN RTX A6000 GPU and Intel Xeon 8358 @ 2.6GHz CPU.


\noindent \textbf{Pre-training and Fine-Tuning Strategies.}\label{sec:curve} 
We first assess the effect of various pre-training protocols (i.e., SSL, SL, and their combinations) and different fine-tuning approaches. For fine-tuning methods,  Frozen, FFT, and our TAMT approaches learn none, full, or partial parameters of the pre-trained model, respectively. Particularly, Frozen method employs a learnable logistic regression classifier for few-shot inference~\cite{10378593}.
As shown in Tab.~\ref{tab:pretrain}, combining SSL with SL as a pre-training protocol consistently yields superior performance\footnote{When a single SSL is used for pre-training, FFT shows a ~2\% improvement over TAMT, likely due to SSL focusing on structural information, while FFT better assimilates semantic information.} over SSL or SL alone, across different fine-tuning methods and various target benchmarks, achieving gains of about 10\%$\sim$20\%.
By integrating SSL with SL for pre-training, our TAMT consistently outperforms FFT by about 1\%$\sim$3\%. These results verify the effectiveness of our TAMT in mitigating the potential risk of overfitting under the optimal SSL \& SL pre-training protocol. Furthermore, as shown in Tab.~\ref{tab:parameter}, our TAMT requires only $\sim$ 0.1× GPU memory, 0.1× learnable parameters, and 0.7× training time compared to FFT, 
indicating the high efficiency of our TAMT decoupled training paradigm.
Additionally, the convergence curves of models trained by FFT and our TAMT on SSV2 dataset~\cite{2017The2} are illustrated in Fig.~\ref{fig:curve}, where we can observe that FFT reaches earlier performance saturation than TAMT at the training stage, and validation performance degrades in the later training period (after $\sim$30 epochs). This may suggest that FFT, with fine-tuning all parameters, suffers from the issue of overfitting again.


\noindent \textbf{Effect of TAA and GTMT.}
We explore the individual contributions of TAA and GTMT within our HTTN in Tab.~\ref{tab:HTTN_components}.
When evaluated separately, both TAA and GTMT exhibit notable performance enhancements, as detailed in Tab.~\ref{tab:HTTN_components} (a).
Specifically, in the absence of any adapter, GTMT outperforms the conventional GAP with an overall accuracy improvement of 9.17\%, showcasing remarkable superiority on the challenging SSV2 and Diving with improvements of 12.96\% and 14.78\%, respectively. 
In addition, by integrating TAA with a variety of global statistical methods, our GTMT surpasses its counterparts, achieving the highest average performance of 66.76\% across three datasets.
It is evident that the approaches utilizing second-order statistics (TCov, ELSTC and GTMT) generally outperform first-order methodology (GAP). And among various second-order fashions, our ELSTC notably exceeds TCov with 0.75\% average 
gains in a more efficient manner (with 262K \textit{vs.} 4K dimension gap, see supplementary material for detailed analysis).
Moreover, by adopting first-order for ELSTC, GTMT achieves a further improvement of 0.82\%. These findings prove that TAA and GTMT can serve as competitive solutions for local and global tuning strategies.
Furthermore, our TAMT also consistently surpasses its adapter counterparts—Adapter~\cite{houlsby2019parameter}, SSF~\cite{Lian_2022_SSF}, TAM~\cite{liu2021tam} and ST-Adapter~\cite{pan2022st}—with average performance improvements of 4.17\%, 7.69\%, 2.97\% and 2.58\%  as detailed in Tab.~\ref{tab:HTTN_components} (b), respectively.
\section{Conclusion}
This paper proposed a novel Temporal-Aware Model Tuning (TAMT) method for cross-domain few-shot action recognition (CDFSAR) task. Particularly, to our best knowledge, our TAMT makes the first attempt to introduce a decoupled training paradigm for CDFSAR, effectively avoiding model retraining in the case of single source data and multiple target data. Moreover, from the perspectives of local feature recalibration and global (powerful) representation generation, a Hierarchical Temporal Tuning Network (HTTN) is proposed to effectively transfer the pre-trained models to target domain in a memory- and parameter-efficient manner. 
Extensive comparisons on CDFSAR tasks verify the effectiveness of our TAMT. 
We believe our TAMT provides a strong baseline for CDFSAR, and potentially contributes to push CDFSAR forward.\\

\section*{Acknowledgement}
This work was supported in part by the National Natural Science Foundation of China under Grant 62276186, 62471083, 61971086, in part by the Haihe Lab of ITAI under Grant 22HHXCJC00002.

{
    \small
    \bibliographystyle{ieeenat_fullname}
    \bibliography{main}

\begin{thebibliography}{60}
\providecommand{\natexlab}[1]{#1}
\providecommand{\url}[1]{\texttt{#1}}
\expandafter\ifx\csname urlstyle\endcsname\relax
  \providecommand{\doi}[1]{doi: #1}\else
  \providecommand{\doi}{doi: \begingroup \urlstyle{rm}\Url}\fi

\bibitem[Alanov et~al.(2023)Alanov, Titov, Nakhodnov, and Vetrov]{Cross_domain_r2}
Aibek Alanov, Vadim Titov, Maksim Nakhodnov, and Dmitry Vetrov.
\newblock Style{D}omain: Efficient and lightweight parameterizations of {StyleGAN} for one-shot and few-shot domain adaptation.
\newblock In \emph{Proceedings of the IEEE/CVF International Conference on Computer Vision (ICCV)}, pages 2184--2194, 2023.

\bibitem[Arnab et~al.(2021)Arnab, Dehghani, Heigold, Sun, Lucic, and Schmid]{arnab2021vivit}
Anurag Arnab, Mostafa Dehghani, Georg Heigold, Chen Sun, Mario Lucic, and Cordelia Schmid.
\newblock V{iViT}: A video vision transformer.
\newblock In \emph{IEEE/CVF International Conference on Computer Vision (ICCV)}, pages 6816--6826, 2021.

\bibitem[Bishay et~al.(2019)Bishay, Zoumpourlis, and Patras]{2019TARN}
Mina Bishay, Georgios Zoumpourlis, and Ioannis Patras.
\newblock T{ARN}: Temporal attentive relation network for few-shot and zero-shot action recognition.
\newblock \emph{British Machine Vision Conference (BMVC)}, pages 130.1--130.14, 2019.

\bibitem[Bishop and Nasrabadi(2006)]{bishop2006pattern}
Christopher~M Bishop and Nasser~M Nasrabadi.
\newblock \emph{Pattern recognition and machine learning}.
\newblock Springer, 2006.

\bibitem[Cao et~al.(2021)Cao, Li, Lv, Wang, and Zhang]{2021Few2}
Congqi Cao, Yajuan Li, Qinyi Lv, Peng Wang, and Yanning Zhang.
\newblock Few-shot action recognition with implicit temporal alignment and pair similarity optimization.
\newblock \emph{Computer Vision and Image Understanding (CVIU)}, 210:\penalty0 103250, 2021.

\bibitem[Cao et~al.(2020)Cao, Ji, Cao, Chang, and Niebles]{2020otam}
Kaidi Cao, Jingwei Ji, Zhangjie Cao, Chien~Yi Chang, and Juan~Carlos Niebles.
\newblock Few-shot video classification via temporal alignment.
\newblock In \emph{IEEE/CVF Conference on Computer Vision and Pattern Recognition (CVPR)}, pages 10615--10624, 2020.

\bibitem[Carreira and Zisserman(2017)]{2017The}
Joao Carreira and Andrew Zisserman.
\newblock Quo vadis, {A}ction {R}ecognition? {A} new model and the kinetics dataset.
\newblock In \emph{proceedings of the IEEE Conference on Computer Vision and Pattern Recognition (CVPR)}, pages 6299--6308, 2017.

\bibitem[Chen et~al.(2023)Chen, Liu, Wang, Zhang, Torr, Zhang, and Tang]{chen2023tem}
Guangyi Chen, Xiao Liu, Guangrun Wang, Kun Zhang, Philip~HS Torr, Xiao-Ping Zhang, and Yansong Tang.
\newblock Tem-adapter: Adapting image-text pretraining for video question answer.
\newblock In \emph{Proceedings of the IEEE/CVF International Conference on Computer Vision (ICCV)}, pages 13945--13955, 2023.

\bibitem[Chen et~al.(2021)Chen, Zou, and Wang]{2021Semantic}
Zixuan Chen, Junhong Zou, and Xiaotao Wang.
\newblock Semantic segmentation on {VSPW} dataset through aggregation of transformer models.
\newblock \emph{arXiv preprint arXiv:2109.01316}, 2021.

\bibitem[Deng et~al.(2009)Deng, Dong, Socher, Li, Li, and Fei-Fei]{ImageNet1k}
Jia Deng, Wei Dong, Richard Socher, Li-Jia Li, Kai Li, and Li Fei-Fei.
\newblock {ImageNet}: A large-scale hierarchical image database.
\newblock In \emph{IEEE Conference on Computer Vision and Pattern Recognition (CVPR)}, pages 248--255, 2009.

\bibitem[Gao et~al.(2023{\natexlab{a}})Gao, Wang, Lin, Zhu, Hu, and Zhou]{gao2023tuning}
Mingze Gao, Qilong Wang, Zhenyi Lin, Pengfei Zhu, Qinghua Hu, and Jingbo Zhou.
\newblock Tuning pre-trained model via moment probing.
\newblock In \emph{Proceedings of the IEEE/CVF International Conference on Computer Vision (ICCV)}, pages 11803--11813, 2023{\natexlab{a}}.

\bibitem[Gao et~al.(2018)Gao, Han, Zhu, Zhang, and Wang]{DBLP:journals/access/GaoHZZW18}
Zan Gao, Tao{-}tao Han, Lei Zhu, Hua Zhang, and Yinglong Wang.
\newblock Exploring the cross-domain action recognition problem by deep feature learning and cross-domain learning.
\newblock \emph{{IEEE} Access}, 6:\penalty0 68989--69008, 2018.

\bibitem[Gao et~al.(2020)Gao, Guo, Guan, Liu, and Chen]{2020A2}
Zan Gao, Leming Guo, Weili Guan, Anan Liu, and Shengyong Chen.
\newblock A pairwise attentive adversarial spatiotemporal network for cross-domain few-shot action recognition-{R}2.
\newblock \emph{{IEEE} Transactions on Image Processing}, 30:\penalty0 767--782, 2020.

\bibitem[Gao et~al.(2022)Gao, Zhao, Zhang, Chen, Liu, and Chen]{DBLP:journals/tcyb/GaoZZCLC22}
Zan Gao, Yibo Zhao, Hua Zhang, Da Chen, An{-}An Liu, and Shengyong Chen.
\newblock A novel multiple-view adversarial learning network for unsupervised domain adaptation action recognition.
\newblock \emph{{IEEE} Transactions on Cybernetics}, 52\penalty0 (12):\penalty0 13197--13211, 2022.

\bibitem[Gao et~al.(2023{\natexlab{b}})Gao, Huang, Zhang, Liu, and Ma]{Cross_domain_r1}
Zhiqiang Gao, Kaizhu Huang, Rui Zhang, Dawei Liu, and Jieming Ma.
\newblock Towards better robustness against common corruptions for unsupervised domain adaptation.
\newblock In \emph{IEEE International Conference on Computer Vision (ICCV)}, pages 18882--18893, 2023{\natexlab{b}}.

\bibitem[Goyal et~al.(2017)Goyal, Kahou, Michalski, Materzynska, Westphal, Kim, Haenel, Fruend, Yianilos, Mueller-Freitag, et~al.]{2017The2}
Raghav Goyal, Samira~Ebrahimi Kahou, Vincent Michalski, Joanna Materzynska, Susanne Westphal, Heuna Kim, Valentin Haenel, Ingo Fruend, Peter Yianilos, Moritz Mueller-Freitag, et~al.
\newblock The “{Something Something}” video database for learning and evaluating visual common sense.
\newblock In \emph{IEEE International Conference on Computer Vision (ICCV)}, pages 5843--5851, 2017.

\bibitem[Guo et~al.(2025{\natexlab{a}})Guo, Wang, Qi, Zhu, and Sun]{2312.01083}
Fei Guo, YiKang Wang, Han Qi, Li Zhu, and Jing Sun.
\newblock Consistency prototype module and motion compensation for few-shot action recognition ({CLIP-CPM}$^2${C}).
\newblock \emph{Neurocomputing}, 611:\penalty0 128649, 2025{\natexlab{a}}.

\bibitem[Guo et~al.(2025{\natexlab{b}})Guo, Wang, Qi, Zhu, and Sun]{guo2024dmsd}
Fei Guo, Yi~Kang Wang, Han Qi, Li Zhu, and Jing Sun.
\newblock {DMSD-CDFSAR}: Distillation from mixed-source domain for cross-domain few-shot action recognition.
\newblock \emph{Expert Systems With Applications}, 270, 2025{\natexlab{b}}.

\bibitem[Guo et~al.(2020)Guo, Codella, Karlinsky, Codella, and Feris]{2020A}
Yunhui Guo, Noel~C. Codella, Leonid Karlinsky, James~V. Codella, and Rogerio Feris.
\newblock A broader study of cross-domain few-shot learning.
\newblock In \emph{European Conference on Computer Vision (ECCV)}, pages 124--141, 2020.

\bibitem[He et~al.(2022)He, Chen, Xie, Li, Doll{\'a}r, and Girshick]{he2022masked}
Kaiming He, Xinlei Chen, Saining Xie, Yanghao Li, Piotr Doll{\'a}r, and Ross Girshick.
\newblock Masked autoencoders are scalable vision learners.
\newblock In \emph{Proceedings of the IEEE/CVF Conference on Computer Vision and Pattern Recognition (CVPR)}, pages 16000--16009, 2022.

\bibitem[Houlsby et~al.(2019)Houlsby, Giurgiu, Jastrzebski, Morrone, De~Laroussilhe, Gesmundo, Attariyan, and Gelly]{houlsby2019parameter}
Neil Houlsby, Andrei Giurgiu, Stanislaw Jastrzebski, Bruna Morrone, Quentin De~Laroussilhe, Andrea Gesmundo, Mona Attariyan, and Sylvain Gelly.
\newblock Parameter-efficient transfer learning for {NLP}.
\newblock In \emph{International Conference on Machine Learning (ICML)}, pages 2790--2799, 2019.

\bibitem[Islam et~al.(2021)Islam, Chen, Panda, Karlinsky, Feris, and Radke]{islam2021dynamic}
Ashraful Islam, Chun-Fu~Richard Chen, Rameswar Panda, Leonid Karlinsky, Rogerio Feris, and Richard~J Radke.
\newblock Dynamic distillation network for cross-domain few-shot recognition with unlabeled data.
\newblock \emph{Advances in Neural Information Processing Systems (NeurIPS)}, 34:\penalty0 3584--3595, 2021.

\bibitem[Kuehne et~al.(2011)Kuehne, Jhuang, Garrote, Poggio, and Serre]{2011HMDB}
H. Kuehne, H. Jhuang, E. Garrote, T. Poggio, and T. Serre.
\newblock H{MDB}: A large video database for human motion recognition.
\newblock In \emph{International Conference on Computer Vision (ICCV)}, pages 2556--2563, 2011.

\bibitem[Li et~al.(2023)Li, Wang, He, Li, Wang, Wang, and Qiao]{uniformerv2}
Kunchang Li, Yali Wang, Yinan He, Yizhuo Li, Yi Wang, Limin Wang, and Yu Qiao.
\newblock Uni{FormerV2}: Unlocking the potential of image vits for video understanding.
\newblock In \emph{Proceedings of the IEEE/CVF International Conference on Computer Vision (ICCV)}, pages 1632--1643, 2023.

\bibitem[Li et~al.(2018)Li, Li, and Vasconcelos]{2018RESOUND}
Yingwei Li, Yi Li, and Nuno Vasconcelos.
\newblock R{ESOUND}: Towards action recognition without representation bias.
\newblock In \emph{Proceedings of the European Conference on Computer Vision (ECCV)}, pages 513--528, 2018.

\bibitem[Lian et~al.(2022)Lian, Zhou, Feng, and Wang]{Lian_2022_SSF}
Dongze Lian, Daquan Zhou, Jiashi Feng, and Xinchao Wang.
\newblock Scaling \& shifting your features: A new baseline for efficient model tuning.
\newblock In \emph{Advances in Neural Information Processing Systems (NeurIPS)}, pages 109--123, 2022.

\bibitem[Lin et~al.(2023)Lin, Wang, He, Chen, Xu, and Zhao]{9970717}
Liqun Lin, Zheng Wang, Jiachen He, Weiling Chen, Yiwen Xu, and Tiesong Zhao.
\newblock Deep quality assessment of compressed videos: A subjective and objective study.
\newblock \emph{IEEE Transactions on Circuits and Systems for Video Technology}, 33\penalty0 (6):\penalty0 2616--2626, 2023.

\bibitem[Liu et~al.(2023)Liu, Zhang, Pirsiavash, and Liu]{Liu_2023_WACV}
Xin Liu, Huanle Zhang, Hamed Pirsiavash, and Xin Liu.
\newblock M{ASTAF}: A model-agnostic spatio-temporal attention fusion network for few-shot video classification.
\newblock In \emph{Proceedings of the IEEE/CVF Winter Conference on Applications of Computer Vision (WACV)}, pages 2508--2517, 2023.

\bibitem[Liu et~al.(2021)Liu, Wang, Wu, Qian, and Lu]{liu2021tam}
Zhaoyang Liu, Limin Wang, Wayne Wu, Chen Qian, and Tong Lu.
\newblock T{AM}: Temporal adaptive module for video recognition.
\newblock In \emph{Proceedings of the IEEE/CVF International Conference on Computer Vision (ICCV)}, pages 13708--13718, 2021.

\bibitem[Ma et~al.(2023)Ma, Fu, Zheng, Peng, Cao, and Huang]{Cross_domain_r3}
Kang Ma, Ying Fu, Dezhi Zheng, Yunjie Peng, Chunshui Cao, and Yongzhen Huang.
\newblock Fine-grained unsupervised domain adaptation for gait recognition.
\newblock In \emph{Proceedings of the IEEE/CVF International Conference on Computer Vision (ICCV)}, pages 11313--11322, 2023.

\bibitem[Miech et~al.(2020)Miech, Alayrac, Laptev, Sivic, and Zisserman]{2020RareAct}
Antoine Miech, Jean~Baptiste Alayrac, Ivan Laptev, Josef Sivic, and Andrew Zisserman.
\newblock Rare{A}ct: A video dataset of unusual interactions.
\newblock \emph{arXiv preprint arXiv:2008.01018}, 2020.

\bibitem[Oquab et~al.(2023)Oquab, Darcet, Moutakanni, Vo, Szafraniec, Khalidov, Fernandez, Haziza, Massa, El-Nouby, et~al.]{DINOV2}
Maxime Oquab, Timoth{\'e}e Darcet, Th{\'e}o Moutakanni, Huy Vo, Marc Szafraniec, Vasil Khalidov, Pierre Fernandez, Daniel Haziza, Francisco Massa, Alaaeldin El-Nouby, et~al.
\newblock D{INO}v2: Learning robust visual features without supervision.
\newblock \emph{arXiv preprint arXiv:2304.07193}, 2023.

\bibitem[Pan et~al.(2022)Pan, Lin, Zhu, Shao, and Li]{pan2022st}
Junting Pan, Ziyi Lin, Xiatian Zhu, Jing Shao, and Hongsheng Li.
\newblock {ST}-{A}dapter: Parameter-efficient image-to-video transfer learning.
\newblock \emph{Advances in Neural Information Processing Systems (ICCV)}, 35:\penalty0 26462--26477, 2022.

\bibitem[Perrett et~al.(2021)Perrett, Masullo, Burghardt, Mirmehdi, and Damen]{2021Temporal}
Toby Perrett, Alessandro Masullo, Tilo Burghardt, Majid Mirmehdi, and Dima Damen.
\newblock Temporal-relational cross {Transformers} for few-shot action recognition.
\newblock In \emph{IEEE/CVF Conference on Computer Vision and Pattern Recognition (CVPR)}, pages 475--484, 2021.

\bibitem[Phoo and Hariharan(2021)]{phoo2020self}
Cheng~Perng Phoo and Bharath Hariharan.
\newblock Self-training for few-shot transfer across extreme task differences.
\newblock In \emph{International Conference on Learning Representations (ICLR)}, 2021.

\bibitem[Piergiovanni et~al.(2023)Piergiovanni, Kuo, and Angelova]{0Rethinking}
A.~J. Piergiovanni, Weicheng Kuo, and Anelia Angelova.
\newblock Rethinking {V}ideo {ViTs}: Sparse video tubes for joint image and video learning.
\newblock In \emph{IEEE/CVF Conference on Computer Vision and Pattern Recognition (CVPR)}, pages 2214--2224, 2023.

\bibitem[Qi et~al.(2023)Qi, Dong, Fan, Ge, Zhang, Ma, and Yi]{qi2023recon}
Zekun Qi, Runpei Dong, Guofan Fan, Zheng Ge, Xiangyu Zhang, Kaisheng Ma, and Li Yi.
\newblock Contrast with reconstruct: Contrastive 3{D} representation learning guided by generative pretraining.
\newblock In \emph{International Conference on Machine Learning (ICML)}, 2023.

\bibitem[Radford et~al.(2021)Radford, Kim, Hallacy, Ramesh, Goh, Agarwal, Sastry, Askell, Mishkin, and Clark]{CLIP_ICML}
Alec Radford, Jong~Wook Kim, Chris Hallacy, Aditya Ramesh, Gabriel Goh, Sandhini Agarwal, Girish Sastry, Amanda Askell, Pamela Mishkin, and Jack Clark.
\newblock Learning transferable visual models from natural language supervision.
\newblock In \emph{International Conference on Machine Learning (ICML)}, pages 8748--8763. PMLR, 2021.

\bibitem[Samarasinghe et~al.(2023)Samarasinghe, Rizve, Kardan, and Shah]{10378593}
Sarinda Samarasinghe, Mamshad~Nayeem Rizve, Navid Kardan, and Mubarak Shah.
\newblock {CDFSL}-{V}: Cross-domain few-shot learning for videos.
\newblock In \emph{Proceedings of the IEEE/CVF international conference on computer vision (ICCV)}, pages 11643--11652, 2023.

\bibitem[Shi et~al.(2024)Shi, Wu, Lin, and Luo]{KP_FSL}
Yuheng Shi, Xinxiao Wu, Hanxi Lin, and Jiebo Luo.
\newblock Commonsense knowledge prompting for few-shot action recognition in videos.
\newblock \emph{IEEE Transactions on Multimedia}, 26:\penalty0 8395--8405, 2024.

\bibitem[Snell et~al.(2017)Snell, Swersky, and Zemel]{snell2017prototypical}
Jake Snell, Kevin Swersky, and Richard Zemel.
\newblock Prototypical networks for few-shot learning.
\newblock In \emph{Proceedings of the 31st International Conference on Neural Information Processing Systems (NeurIPS)}, pages 4080--4090, 2017.

\bibitem[Soomro et~al.(2012)Soomro, Roshan~Zamir, and Shah]{2012UCF101}
Khurram Soomro, Amir Roshan~Zamir, and Mubarak Shah.
\newblock {UCF101}: A dataset of 101 human actions classes from videos in the wild.
\newblock \emph{arXiv preprint arXiv:1212.0402}, 2012.

\bibitem[Thatipelli et~al.(2022)Thatipelli, Narayan, Khan, Anwer, Khan, and Ghanem]{2021Spatio}
Anirudh Thatipelli, Sanath Narayan, Salman Khan, Rao~Muhammad Anwer, Fahad~Shahbaz Khan, and Bernard Ghanem.
\newblock Spatio-temporal relation modeling for few-shot action recognition.
\newblock In \emph{IEEE/CVF Conference on Computer Vision and Pattern Recognition (CVPR)}, pages 19926--19935, 2022.

\bibitem[Tong et~al.(2022)Tong, Song, Wang, and Wang]{2022VideoMAE}
Zhan Tong, Yibing Song, Jue Wang, and Limin Wang.
\newblock {VideoMAE}: Masked autoencoders are data-efficient learners for self-supervised video pre-training.
\newblock \emph{Advances in neural information processing systems (NeurIPS)}, 35:\penalty0 10078--10093, 2022.

\bibitem[Wang et~al.(2023{\natexlab{a}})Wang, Huang, Zhao, Tong, He, Wang, Wang, and Qiao]{videomaev2}
Limin Wang, Bingkun Huang, Zhiyu Zhao, Zhan Tong, Yinan He, Yi Wang, Yali Wang, and Yu Qiao.
\newblock Video{MAE} {V}2: Scaling video masked autoencoders with dual masking.
\newblock In \emph{Proceedings of the IEEE/CVF Conference on Computer Vision and Pattern Recognition (CVPR)}, pages 14549--14560, 2023{\natexlab{a}}.

\bibitem[Wang et~al.(2022{\natexlab{a}})Wang, Zhang, Qing, Tang, Zuo, Gao, Jin, and Sang]{2022Hybrid}
Xiang Wang, Shiwei Zhang, Zhiwu Qing, Mingqian Tang, Zhengrong Zuo, Changxin Gao, Rong Jin, and Nong Sang.
\newblock Hybrid relation guided set matching for few-shot action recognition.
\newblock In \emph{IEEE/CVF Conference on Computer Vision and Pattern Recognition (CVPR)}, pages 19916--19925, 2022{\natexlab{a}}.

\bibitem[Wang et~al.(2023{\natexlab{b}})Wang, Wang, Jiang, and Luo]{wang2023few2}
Xixi Wang, Xiao Wang, Bo Jiang, and Bin Luo.
\newblock Few-shot learning meets transformer: Unified query-support transformers for few-shot classification.
\newblock \emph{IEEE Transactions on Circuits and Systems for Video Technology}, 33\penalty0 (12):\penalty0 7789--7802, 2023{\natexlab{b}}.

\bibitem[Wang et~al.(2023{\natexlab{c}})Wang, Ye, Qi, Wang, Wu, Shan, Qie, and Wang]{wang2023task}
Xiao Wang, Weirong Ye, Zhongang Qi, Guangge Wang, Jianping Wu, Ying Shan, Xiaohu Qie, and Hanzi Wang.
\newblock Task-aware dual-representation network for few-shot action recognition.
\newblock \emph{IEEE Transactions on Circuits and Systems for Video Technology}, 33\penalty0 (10):\penalty0 5932--5946, 2023{\natexlab{c}}.

\bibitem[Wang et~al.(2023{\natexlab{d}})Wang, Zhang, Qing, Gao, Zhang, Zhao, and Sang]{wang2023molo}
Xiang Wang, Shiwei Zhang, Zhiwu Qing, Changxin Gao, Yingya Zhang, Deli Zhao, and Nong Sang.
\newblock Mo{L}o: Motion-augmented long-short contrastive learning for few-shot action recognition.
\newblock In \emph{Proceedings of the IEEE/CVF Conference on Computer Vision and Pattern Recognition (CVPR)}, pages 18011--18021, 2023{\natexlab{d}}.

\bibitem[Wang et~al.(2023{\natexlab{e}})Wang, Zhang, Qing, Lv, Gao, and Sang]{WANG2023103737}
Xiang Wang, Shiwei Zhang, Zhiwu Qing, Yiliang Lv, Changxin Gao, and Nong Sang.
\newblock Cross-domain few-shot action recognition with unlabeled videos.
\newblock \emph{Computer Vision and Image Understanding (CVIU)}, 233:\penalty0 103737, 2023{\natexlab{e}}.

\bibitem[Wang et~al.(2023{\natexlab{f}})Wang, Zhang, Yuan, Zhang, Gao, Zhao, and Sang]{wang2023few}
Xiang Wang, Shiwei Zhang, Hangjie Yuan, Yingya Zhang, Changxin Gao, Deli Zhao, and Nong Sang.
\newblock Few-shot action recognition with captioning foundation models.
\newblock \emph{arXiv preprint arXiv:2310.10125}, 2023{\natexlab{f}}.

\bibitem[Wang et~al.(2024{\natexlab{a}})Wang, Lu, Yu, Pang, and Wang]{10489992}
Xiao Wang, Yang Lu, Wanchuan Yu, Yanwei Pang, and Hanzi Wang.
\newblock Few-shot action recognition via multi-view representation learning.
\newblock \emph{IEEE Transactions on Circuits and Systems for Video Technology}, pages 1--1, 2024{\natexlab{a}}.

\bibitem[Wang et~al.(2024{\natexlab{b}})Wang, Zhang, Cen, Gao, Zhang, Zhao, and Sang]{wang2023clip}
Xiang Wang, Shiwei Zhang, Jun Cen, Changxin Gao, Yingya Zhang, Deli Zhao, and Nong Sang.
\newblock {CLIP}-guided prototype modulating for few-shot action recognition.
\newblock \emph{International Journal of Computer Vision}, 132\penalty0 (6):\penalty0 1899--1912, 2024{\natexlab{b}}.

\bibitem[Wang et~al.(2022{\natexlab{b}})Wang, Li, Li, He, Huang, Zhao, Zhang, Xu, Liu, Wang, Xing, Chen, Pan, Yu, Wang, Wang, and Qiao]{wang2022internvideo}
Yi Wang, Kunchang Li, Yizhuo Li, Yinan He, Bingkun Huang, Zhiyu Zhao, Hongjie Zhang, Jilan Xu, Yi Liu, Zun Wang, Sen Xing, Guo Chen, Junting Pan, Jiashuo Yu, Yali Wang, Limin Wang, and Yu Qiao.
\newblock {InternVideo}: General video foundation models via generative and discriminative learning.
\newblock \emph{arXiv preprint arXiv:2212.03191}, 2022{\natexlab{b}}.

\bibitem[Wanyan et~al.(2023)Wanyan, Yang, Chen, and Xu]{AMFTR_CVPR23}
Yuyang Wanyan, Xiaoshan Yang, Chaofan Chen, and Changsheng Xu.
\newblock Active exploration of multimodal complementarity for few-shot action recognition.
\newblock In \emph{Proceedings of the IEEE/CVF Conference on Computer Vision and Pattern Recognition (CVPR)}, pages 6492--6502, 2023.

\bibitem[Xing et~al.(2023)Xing, Wang, Ruan, Chen, Guo, Mu, Dai, Wang, and Liu]{GgHM_ICCV23}
Jiazheng Xing, Mengmeng Wang, Yudi Ruan, Bofan Chen, Yaowei Guo, Boyu Mu, Guang Dai, Jingdong Wang, and Yong Liu.
\newblock Boosting few-shot action recognition with graph-guided hybrid matching.
\newblock In \emph{Proceedings of the IEEE/CVF International Conference on Computer Vision (ICCV)}, pages 1740--1750, 2023.

\bibitem[Yan et~al.(2022)Yan, Xiong, Arnab, Lu, Zhang, Sun, and Schmid]{2022Multiview}
Shen Yan, Xuehan Xiong, Anurag Arnab, Zhichao Lu, Mi Zhang, Chen Sun, and Cordelia Schmid.
\newblock Multiview {Transformers} for video recognition.
\newblock In \emph{IEEE/CVF Conference on Computer Vision and Pattern Recognition (CVPR)}, pages 3323--3333, 2022.

\bibitem[Zhang et~al.(2021)Zhang, Zhou, and He]{2021Learning}
Songyang Zhang, Jiale Zhou, and Xuming He.
\newblock Learning implicit temporal alignment for few-shot video classification.
\newblock \emph{International Joint Conference on Artificial Intelligence (IJCAI)}, pages 1309--1315, 2021.

\bibitem[Zhou et~al.(2021)Zhou, Wei, Wang, Shen, Xie, Yuille, and Kong]{iBoT_ICLR21}
Jinghao Zhou, Chen Wei, Huiyu Wang, Wei Shen, Cihang Xie, Alan Yuille, and Tao Kong.
\newblock Image {BERT} pre-training with online tokenizer.
\newblock In \emph{International Conference on Learning Representations (ICLR)}, 2021.

\bibitem[Zhu and Yang(2018)]{2018Compound}
Linchao Zhu and Yi Yang.
\newblock Compound memory networks for few-shot video classification.
\newblock In \emph{Proceedings of the European Conference on Computer Vision (ECCV)}, pages 751--766, 2018.

\end{thebibliography}
}
\newpage
\renewcommand*{\thetable}{S\arabic{table}}
\renewcommand*{\thefigure}{S\arabic{figure}}
\setcounter{page}{1}
\maketitlesupplementary
\newcommand{\fisota}[1]{\textcolor{red}{\textbf{#1}}}
\newcommand{\sesota}[1]{\textcolor{blue}{\textbf{#1}}}

In the supplementary materials, we first explore the effect of hyper-parameters on the Hierarchical Temporal Tuning Network (HTTN), mainly including the number $L$ and parameters $\gamma$ \& $\beta$ for TAA blocks as well as the number of group $G$ for ELSTC. Furthermore, to fully evaluate the generalization of TAMT, we set up a setting called generalization across datasets, and finally show its performance of varying shots and FSAR tasks. Lastly, we conduct visualization analyses to further validate the effectiveness of our TAMT.

\section{Effect of Hyper-parameters on HTTN}
\noindent \textbf{Number $L$ and Parameters $\gamma$ \& $\beta$ in TAA block.} 
In Tab.~\ref{tab:HTTNGL} and Tab.~\ref{tab:9}, we explore the optimal TAA transformer block number $L$ and the parameter sharing strategy for parameters $\gamma$ and $\beta$. Initially, $L$ varies from 0 to 4, among which an $L$ value of 0 implies a configuration without any adapters. Relative to this baseline ($L=0$), the introduction of adapters yields a positive impact, enhancing performance by over 1.96\% with only a minimal increase in training cost.
Optimal performance is observed when $L$ is set to 2 or 3. For higher efficiency and considering the performance-consumption balance, $L=2$ is chosen as the default configuration. For the parameters $\gamma$ and $\beta$ in TAA blocks, they are partially shared. Only the $\mathbf{W}_{\downarrow}^{\gamma}$ and $\mathbf{W}_{\downarrow}^{\beta}$ in Eqn.~(2) \& Eqn.~(3) are shared, resulting in an average gain of 0.48\% across five datasets while reducing learnable parameters by 10\%, as shown in Tab.~\ref{tab:9}.

\noindent \textbf{Number of Group G for ELSTC.} To evaluate the effect of different group numbers $G$ in ELSTC, we consider values ranging from 1 to 8 for a sequential feature of length $T=8$. We observe both performance and computational overhead (including feature dimension, number of parameters, GFLOPs, and inference latency), as shown in Tab.~\ref{tab:group-number}. The results show that grouping features ($G>1$) effectively reduces the computational overhead compared to the original setting ($G=1$). Moreover, increasing $G$ further alleviates the overhead. Notably, optimal performance is achieved at $G=4$, 
This improvement likely results from a balance between more effective optimization (compared with $G=1, 2$) and better preservation of temporal interactions within each group (in contrast to $G=8$).

\begin{table}[t]
    \centering
    \fontsize{9}{9.6}\selectfont
    \setlength{\tabcolsep}{6pt}
    \renewcommand{\arraystretch}{1.5}
    \centering

                \begin{tabular}{c|ccc|cc}\hline
                  \hspace{1mm}  $L$ \hspace{1mm} & SSV2 & Diving & UCF & Average & Memory\\ \hline
                  \hspace{1mm}  0 \hspace{1mm} & 53.41 & 42.87 & 94.97 & 63.58 & 1.2G\\
                    1 & 57.48 & 43.52 & 95.61  & 65.54& +0.0G\\
                    2 &  \hlours{59.18} & \hlours{45.18} & \hlours{95.92}  & \hlours{66.76}& +0.7G\\
                    3  & 59.86 & 44.75 & 95.70 & 66.77 & +1.8G\\
                    4 & 59.67 & 44.22 & 94.87 & 66.25 & +2.9G\\
                    \hline
                \end{tabular}
    \caption{Effect of the hyper-parameter $L$ on HTTN, and the accuracy (\%) of 5-way 5-shot is reported. Memory: GPU memory for training.}
    \label{tab:HTTNGL}
\end{table}

\begin{table}[!ht]
\centering
\fontsize{9}{9.6}\selectfont
\setlength{\tabcolsep}{3.5pt}
\renewcommand{\arraystretch}{1.5}
\begin{tabular}{c|c|ccccc|c}
\hline
 & P & HMDB & SSV2 & Diving & UCF & RareAct & Average 
 \\ \hline
S & 2.8M & 74.14 & 59.18 & 45.18 & 95.92 & 67.44 & 68.37 \\
U & 3.1M & 73.94 & 58.34 & 44.12 & 95.58 & 67.47 & 67.89 \\ \hline
\end{tabular}
\caption{Comparison (\%) of the shared parameters $\gamma \& \beta$ in $\mathbf{W}_{\downarrow}^{\gamma \& \beta}$. S: $\gamma \& \beta$ are Shared, U: $\gamma \& \beta$ are not Shared. P: Parameters.}
\label{tab:9}
\end{table}

\begin{table*}[!ht]
\centering
\fontsize{9}{8.6}\selectfont
\setlength{\tabcolsep}{11.8pt}
\renewcommand{\arraystretch}{1.5}
\begin{tabular}{c|ccccc|ccc:c}
\hline
\parbox[c][.9cm][c]{1.1cm}{Feature \\ Splitting} & $G$ & Dim.  & Params. & GFLOPs& Latency & SSv2 & Diving & UCF & Avg.
 \\ \hline
$\times$ & 1 &262K & 101M & 10.5G  & 10.6ms & 59.06 & 43.76 & 95.32 & 66.05 \\ 
\hdashline
\multirow{3}{*}{$\checkmark$} & 2 & 65K & 25M & 4.2G  & 5.5ms & 58.85 & 43.95 & 95.15 & 65.98 \\
& 4  & 4K  & 1.6M & 2.2G & 3.7ms & \hlours{59.18} & \hlours{45.18} & \hlours{95.92} & \hlours{66.76} \\
& 8 & 1K  & 1.0M & 2.1G & 2.8ms & 58.09 & 44.45 & 95.24 & 65.92 \\
\hline
\end{tabular}
\vspace{-7pt}
\caption{Effect of the hyper-parameter $G$ on HTTN, where 5-way 5-shot accuracy (\%) and computation overhead are reported. Dim.: Dimension of $\mathbf{M}_2$. Params.: Training parameters. GFLOPs: GFLOPs of ELSTC. Latency: Inference latency of ELSTC.}
\label{tab:group-number}
\end{table*}

\begin{table*}[!th]
\centering
\fontsize{9}{8.6}\selectfont
\setlength{\tabcolsep}{11pt}
\renewcommand{\arraystretch}{1.5}
\begin{tabular}{c|c|ccccc}
\hline
 \multirow{2}*{Method} &\multirow{2}*{Pre-trained Dataset$~\rightarrow~$Tuned Dataset$~\rightarrow$} &  \multicolumn{5}{c}{Test Dataset}
 \\ \cline{3-7}
 & & HMDB & SSV2 & Diving & UCF & Average 
 \\ \hline
CDFSL-V~\cite{10378593} & \multirow{2}*{K400$~\rightarrow~$HMDB~$\rightarrow$} & - & 21.39 & 21.21 & 51.66 & 31.42 \\
TAMT~(Ours) &  & - & 43.22 & 29.04 & 63.49 & 45.25\textsubscript{(+13.83)} \\
\hdashline
CDFSL-V~\cite{10378593} & \multirow{2}*{K400$~\rightarrow~$UCF~$\rightarrow$} & 51.97 & 24.36 & 22.62 & - & 32.98 \\
TAMT~(Ours) & & 72.50 & 43.02 & 29.21 & - & 48.24\textsubscript{(+15.26)} \\
\hline
\end{tabular}
\caption{Comparison (\%) with CDFSL-V~\cite{10378593} on across datasets setting. All results are conducted on ViT-S network with $112\times 112$ resolution, reported 5-way 5-shot accuracy on test dataset. }
\label{tab:8}
\end{table*}
\begin{table*}[ht!]
    \centering
    \fontsize{9}{8.6}\selectfont
    \setlength{\tabcolsep}{14.25pt}
    \renewcommand{\arraystretch}{1.5}
    \begin{tabular}{l|c|c|c|c|c|c|c}\hline
    \multirow{2}*{Method} & \multirow{2}*{$K$-shot} & \multicolumn{6}{c}{Target} \\
    \cline{3-8}
       &  & HMDB & SSV2 & Diving & UCF & RareAct & Average \\ \hline
      \textcolor{gray}{STARTUP++~\cite{phoo2020self}} & \multirow{4}*{1-shot} & \textcolor{gray}{16.66} & \textcolor{gray}{14.17} & \textcolor{gray}{13.13} & \textcolor{gray}{24.48} & \textcolor{gray}{17.21} & \textcolor{gray}{17.13}\\
      \textcolor{gray}{DD++~\cite{islam2021dynamic}} &  & \textcolor{gray}{17.44} & \textcolor{gray}{14.96} & \textcolor{gray}{13.73} & \textcolor{gray}{26.04} & \textcolor{gray}{19.02} & \textcolor{gray}{18.24} \\
      CDFSL-V~\cite{10378593} &  & \textcolor{gray}{\sesota{18.59}} & \textcolor{gray}{\sesota{16.01}} & \textcolor{gray}{\sesota{14.11}} & \textcolor{gray}{\sesota{27.78}} & \textcolor{gray}{\sesota{20.06}} & \textcolor{gray}{\sesota{19.31}}\\
      TAMT~(Ours) &  & $\textcolor{red}{\textbf{47.02}}$ & $\textcolor{red}{\textbf{34.45}}$ & $\textcolor{red}{\textbf{27.04}}$ & $\textcolor{red}{\textbf{72.38}}$ & $\textcolor{red}{\textbf{36.04}}$ & $\textcolor{red}{\textbf{43.39\textsubscript{(+24.08)}}}$\\
      \hline
      \textcolor{gray}{STARTUP++~\cite{phoo2020self}} & \multirow{6}*{5-shot} & \textcolor{gray}{24.97} & \textcolor{gray}{15.16} & \textcolor{gray}{14.55} & \textcolor{gray}{32.20} & \textcolor{gray}{31.77} & \textcolor{gray}{23.73}\\
      \textcolor{gray}{DD++~\cite{islam2021dynamic}} &  & \textcolor{gray}{25.99} & \textcolor{gray}{16.00} & \textcolor{gray}{16.24} & \textcolor{gray}{34.10} & \textcolor{gray}{31.20} & \textcolor{gray}{24.71}\\
      SEEN*\dag~\cite{WANG2023103737} &  & 52.80  &  31.20  & 40.90 &  79.60 & 50.20 & 50.94\\
      CDFSL-V~\cite{10378593} &  & 29.80 & 17.21 & 16.37 & 36.53& 33.91 & 26.76\\
      DMSD*\dag~\cite{guo2024dmsd} & & \textcolor{blue}{\textbf{54.90}}  & \textcolor{blue}{\textbf{32.10}}  & \textcolor{blue}{\textbf{42.28}} & \textcolor{blue}{\textbf{81.90}} & \textcolor{blue}{\textbf{53.30}} & \textcolor{blue}{\textbf{52.90}}\\
      TAMT~(Ours) & & $\textcolor{red}{\textbf{61.76}}$ & $\textcolor{red}{\textbf{48.90}}$ & $\textcolor{red}{\textbf{38.33}}$ & $\textcolor{red}{\textbf{87.76}}$ & $\textcolor{red}{\textbf{52.81}}$ & $\textcolor{red}{\textbf{57.91\textsubscript{(+31.15)}}}$\\
      \hline
      \textcolor{gray}{STARTUP++~\cite{phoo2020self}} & \multirow{4}*{20-shot}  & \textcolor{gray}{30.48} & \textcolor{gray}{17.15} & \textcolor{gray}{17.30} & \textcolor{gray}{34.02} & \textcolor{gray}{38.45} & \textcolor{gray}{27.48}\\
      \textcolor{gray}{DD++~\cite{islam2021dynamic}} &  & \textcolor{gray}{33.09} & \textcolor{gray}{17.56} & \textcolor{gray}{17.33} & \textcolor{gray}{36.72} & \textcolor{gray}{39.97} & \textcolor{gray}{28.93} \\
      CDFSL-V~\cite{10378593} &  & \textcolor{gray}{\sesota{36.89}} & \textcolor{gray}{\sesota{18.72}} & \textcolor{gray}{\sesota{17.81}} & \textcolor{gray}{\sesota{39.92}} & \textcolor{gray}{\sesota{42.51}} & \textcolor{gray}{\sesota{31.17}}\\
      TAMT~(Ours) &  & $\textcolor{red}{\textbf{73.71}}$ & $\textcolor{red}{\textbf{55.45}}$ & $\textcolor{red}{\textcolor{red}{\textbf{42.68}}}$ & $\textcolor{red}{\textbf{91.38}}$  & ${\textcolor{red}{\textbf{63.27}}}$ & ${\textcolor{red}{\textbf{65.30\textsubscript{(+34.13)}}}}$\\
   \hline
    \end{tabular}
 \caption{Comparison (\%) of state-of-the-arts on various 5-way $K$-shot settings ($K=1,5,20$) of CDFSAR with employing K-100 as source dataset. All results are conducted with $112\times 112$ resolution by using ViT-S backbone, except Method marked by * ($224\times 224$ resolution by using ResNet-18).}
    \label{tab:shots}
\end{table*}

\begin{table*}[h!]
  \centering
  \fontsize{9}{8.6}\selectfont
  \renewcommand{\arraystretch}{1.5}
  \setlength{\tabcolsep}{9pt}
  \scalebox{1.07}{
  \begin{tabular}{l|c|c|c|c|c|c|c}
  \hline
  Method & M. & Pre-training & Tuning & HMDB & SSV2 &UCF & Average\\
  \hline 
    CLIP*~\cite{CLIP_ICML} & \multirow{4}{*}{\rotatebox{90}{Multi-modal}}  & CLIP-ViT-B & Frozen & 58.2/77.0 & 30.0/42.4 & 89.7/95.7& 59.3/71.7\\
    CapFSAR~\cite{wang2023few}  & & BLIP-ViT-B & FFT & 70.3/81.3 & 54.0/70.1 &93.1/97.7& 72.5/83.0\\
    CLIP-CPM$^2$C~\cite{2312.01083}  & & CLIP-ViT-B & FFT & \textcolor{blue}{\textbf{75.9}}/\textcolor{blue}{\textbf{88.0}} & 60.1/\textcolor{blue}{\textbf{72.8}} &  95.0/98.6& 77.0/\textcolor{blue}{\textbf{86.5}} \\
    CLIP-FSAR~\cite{wang2023clip}& & CLIP-ViT-B & FFT & 75.8/87.7 & \textcolor{red}{\textbf{61.9}}/72.1 &  \textcolor{blue}{\textbf{96.6}}/\textcolor{red}{\textbf{99.0}}& \textcolor{blue}{\textbf{78.1}}/86.3\\
    \hdashline
    OTAM*~\cite{2020otam}  &\multirow{5}{*}{\rotatebox{90}{Unimodal}} & CLIP-ViT-B(V) & FFT & 72.5/83.9 & 50.2/68.6 & 95.8/98.8& 72.8/83.8\\
    TRX*~\cite{2021Temporal} & & BLIP-ViT-B(V) & FFT &  58.9/79.9 & 45.1/68.5 & 90.9/97.4 & 65.0/81.9 \\
    HyRSM*~\cite{2022Hybrid}  & & BLIP-ViT-B(V) & FFT &  69.8/80.6 & 52.1/69.5 & 91.6/96.9 & 71.2/82.3 \\
    MASTAF~\cite{Liu_2023_WACV} & & JFT-ViT-B & FFT & 69.5/\textit{N/A}\ \  \ & 60.7/\textit{N/A}\ \  \ &91.6/\textit{N/A}\ \ \ &73.9/\textit{N/A}\ \ \  \\
    TAMT (Ours) &  & ViT-B & PEFT &\textcolor{red}{\textbf{77.7}}/\textcolor{red}{\textbf{88.2}} & \textcolor{blue}{\textbf{61.4}}/\textcolor{red}{\textbf{73.3}}& \textcolor{red}{\textbf{97.5}}/\textcolor{blue}{\textbf{98.8}} & \textcolor{red}{\textbf{78.9}}/\textcolor{red}{\textbf{86.8}}\\ \hline
\end{tabular}
}
\caption{Comparison (\%) of state-of-the-arts on FSAR setting in terms of 5-way 1-shot/5-shot accuracy. M.: Modality, (V): Only visual encoder of CLIP. 
*: from~\cite{wang2023clip,wang2023few}. }
\label{tab:fsar}
\end{table*}

\section{Generalization Verification}
To further validate the generalization of TAMT, we first conduct experiments under a setting of generalization across datasets. Furthermore, we validate the effect of our TAMT on more shot experiments, and finally demonstrate the generalization ability of our proposed HTTN in FSAR tasks.

\noindent \textbf{Generalization Across Datasets.}
Here, we compare with the counterpart CDFSL-V on a challenging setting, where we pre-train the models on the K-400 dataset and fine-tune the models on on UCF or HMDB. Then, the fine-tuned models are directly adopted to four downstream datasets without any tuning. As shown in Tab.~\ref{tab:8},
our TAMT outperforms CDFSL-V by an average of 13.83\% and 15.26\% on four test datasets~\cite{2018RESOUND, 2011HMDB, 2017The2, 2012UCF101}, respectively. These results clearly demonstrate that our method can be well generalized across different datasets.

\noindent \textbf{Results of Different Training Shots.}\label{sec:shots}
To further assess the generalization of our TAMT method, we compare our TAMT on various 5-way $K$-shot ($K=1, 5, 20$) settings, by using ViT-S with $112\times 112$ input resolution. The performance of transferring from source dataset K-100~\cite{2018Compound} to five target datasets~\cite{2018RESOUND, 2011HMDB, 2017The2, 2020RareAct, 2012UCF101} is presented in Tab.~\ref{tab:shots}, in which the average Top-2 best performances are marked by \fisota{red} and \sesota{blue}, respectively.
As shown in Tab.~\ref{tab:shots}, our TAMT exhibits outstanding performance compared to the prime counterpart CDFSL-V~\cite{10378593} under 1-shot, 5-shot and 20-shot settings with a significant margin on average accuracy over 24.08\%, 31.15\% and 34.13\%. Particularly, for the 5-way 1-shot setting, our TAMT is the only approach to achieve a significative performance (namely, above 20\% for 5-way recognition) on HMDB, SSV2, Diving and UCF datasets.
In addition, the performance of TAMT is boosted by 14.52\% and 21.91\%, when extending 1-shot to 5-shot and 20-shot settings, which is more remarkable than the 7.45\% and 11.86\% increase observed in CDFSL-V. All the above results reveal that 
our TAMT has a good ability to explore information lying in the annotated support set, effectively handling the challenging 1-shot setting and benefiting from the increase in support samples.

\noindent \textbf{Generalization on FSAR Task.} 
Our TAMT approach is also evaluated on the conventional FSAR problem, where we compare it alongside very recent Method based on large-scale models, such as CLIP-ViT-B and BLIP-ViT-B, as detailed in Tab.~\ref{tab:fsar}. 
Considering CLIP network is conducted pre-training using 400M data, our TAMT employs the Kinetics-710~\cite{uniformerv2} database for the pre-training phase with about 660K trainable instances.
The results demonstrate that TAMT exhibits impressive performance superiority in dual modality settings. Specifically, TAMT outperforms comparable unimodal competitors by clear margin, which achieves about 5.0\% and 3.0\% on average across multiple datasets. Moreover, TAMT shows performance gain of 0.8\% over huge-pretrained models within the CLIP family in terms of 1-shot accuracy, despite the absence of auxiliary text modality. 
Notably, by using the PEFT training protocol, TAMT theoretically benefits from a lower training complexity than these full fine-tuning (FFT) approaches. 
These results show that TAMT generalizes well to the FSAR setting, providing an efficient and effective alternative.

\begin{figure*}[!th]
  \centering
  \captionsetup[subfigure]{labelformat=empty}

  \begin{subfigure}{\linewidth}
    \centering
    \includegraphics[height=2.1cm,width=15cm]{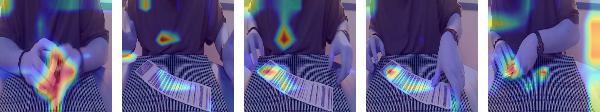}
    \caption{(1a) K-400 dataset, \textit{``Playing cards"}, CDFSL-V~\cite{10378593}}
    \label{fig:vis_source_cdfsl-v}
  \end{subfigure}
  \begin{subfigure}{\linewidth}
    \centering
    \includegraphics[height=2.1cm,width=15cm]{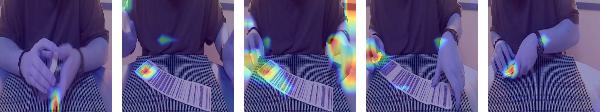}
    \caption{(1b) K-400 dataset, \textit{``Playing cards"}, TAMT (Ours)}
    \label{fig:vis_source_tamt}
  \end{subfigure}
  \begin{subfigure}{\linewidth}
    \centering
    \includegraphics[height=2.1cm,width=15cm]{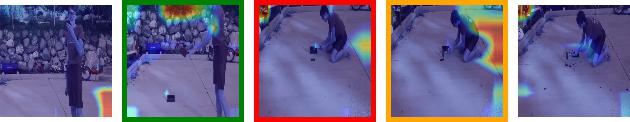}
    \caption{(2a) K-400 $\rightarrow$ RareAct, \textit{``Hammer phone"}, CDFSL-V~\cite{10378593}}
    \label{fig:vis_target_cdfsl-v}
  \end{subfigure}
  \begin{subfigure}{\linewidth}
    \centering
    \includegraphics[height=2.1cm,width=15cm]{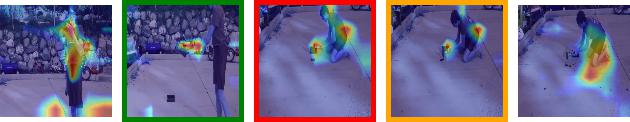}
    \caption{(2b) K-400 $\rightarrow$ RareAct, \textit{``Hammer phone"}, TAMT (Ours)}
    \label{fig:vis_target_tamt}
  \end{subfigure}
  
  \caption{Feature visualization on setting of K-400 $\rightarrow$ RareAct.}
  \label{fig:vis}
\end{figure*}


\section{Visualization Analyses}
To further validate the effectiveness of our TAMT method for addressing the problem of domain gap, we visualize feature heatmaps (the last layer of the backbone) of different models pre-trained on the source dataset (K-400) and those after tuning on the target dataset (RareAct) in Fig.~\ref{fig:vis}. It can be observed that, on the source dataset, both CDFSL-V~\cite{10378593} and our TAMT focus on discriminative regions. After tuning on the target dataset, TAMT captures more semantic features for better recognition (e.g., the human body and phone in class ``\textit{Hammer phone}"), indicating superiority to address the problem of domain gap on downstream tasks.


\end{document}